%% file: sample-sigconf.tex
\documentclass[sigconf]{acmart}

\AtBeginDocument{%
  }

\copyrightyear{2025}
\acmYear{2025}
\setcopyright{acmlicensed}
\acmConference[KDD '25] {Proceedings of the 31st ACM SIGKDD Conference on Knowledge Discovery and Data Mining V.2}{August 3--7, 2025}{Toronto, ON, Canada.}
\acmBooktitle{Proceedings of the 31st ACM SIGKDD Conference on Knowledge Discovery and Data Mining V.2 (KDD '25), August 3--7, 2025, Toronto, ON, Canada}
\acmISBN{979-8-4007-1454-2/25/08}
\acmDOI{10.1145/3711896.3737138}

\definecolor{myred}{RGB}{185,83,83}
\definecolor{myblue}{RGB}{45,68,172}
\definecolor{mygreen}{RGB}{22,164,35}

\usepackage{adjustbox}
\usepackage{multicol}
\usepackage{multirow}
\usepackage{pifont}
\usepackage{ulem}
\usepackage{enumitem}

\begin{document}

\title{Structure-Enhanced Protein Instruction Tuning: \\ Towards General-Purpose Protein Understanding with LLMs}

\author{Wei Wu}
\authornotemark[2]
\orcid{0009-0009-1590-601X}
\affiliation{%
  \institution{School of Artificial Intelligence and Data Science, University of Science and Technology of China}
  \city{Hefei}
  \country{China}
}
\email{urara@mail.ustc.edu.cn}

\author{Chao Wang}
\authornote{Corresponding author.}
\orcid{0000-0001-7717-447X}
\affiliation{%
  \institution{School of Artificial Intelligence and Data Science, University of Science and Technology of China}
  \city{Hefei}
  \country{China}
}
\email{wangchaoai@ustc.edu.cn}

\author{Liyi Chen}
\authornote{Work done during the internship at Alibaba Cloud Computing.}
\orcid{0000-0003-2166-4386}
\affiliation{%
  \institution{School of Artificial Intelligence and Data Science, University of Science and Technology of China}
  \city{Hefei}
  \country{China}
}
\email{liyichen@ustc.edu.cn}

\author{Mingze Yin}
\orcid{0009-0009-6595-9849}
\authornotemark[2]
\affiliation{%
  \institution{College of Computer Science and Technology, Zhejiang University}
  \city{Hangzhou}
  \country{China}
}
\email{12521039@zju.edu.cn}

\author{Yiheng Zhu}
\orcid{0000-0001-8020-9979}
\authornotemark[2]
\affiliation{%
  \institution{College of Computer Science and Technology, Zhejiang University}
  \city{Hangzhou}
  \country{China}
}
\email{zhuyiheng2020@zju.edu.cn}

\author{Kun Fu}
\orcid{0000-0002-2305-1017}
\affiliation{%
  \institution{Alibaba Cloud Computing}
  \city{Beijing}
  \country{China}
}
\email{fukun.fu@alibaba-inc.com}

\author{Jieping Ye}
\orcid{0000-0001-8662-5818}
\affiliation{%
  \institution{Alibaba Cloud Computing}
  \city{Hangzhou}
  \country{China}
}
\email{yejieping.ye@alibaba-inc.com}

\author{Hui Xiong}
\orcid{0000-0001-6016-6465}
\affiliation{%
  \institution{Thrust of Artificial Intelligence, The Hong Kong University of Science and Technology (Guangzhou)}
  \city{Guangzhou}
  \country{China}
}
\affiliation{%
  \institution{Department of Computer Science and Engineering, The Hong Kong University of Science and Technology}
  \city{Hong Kong SAR}
  \country{China}
}
\email{xionghui@ust.hk}

\author{Zheng Wang}
\orcid{0009-0008-4271-6206}
\affiliation{%
  \institution{Alibaba Cloud Computing}
  \city{Beijing}
  \country{China}
}
\email{wz388779@alibaba-inc.com}

\renewcommand{\shortauthors}{Wei Wu et al.}


\input{sections/abstract}




\maketitle
\newcommand\kddavailabilityurl{https://doi.org/10.5281/zenodo.15510090}

\ifdefempty{\kddavailabilityurl}{}{
\begingroup\small\noindent\raggedright\textbf{KDD Availability Link:}\\
The code, datasets and model checkpoints of this paper has been made publicly available at \url{\kddavailabilityurl}.
\endgroup
}

\input{sections/introduction}
\input{sections/relatedwork}
\input{sections/dataset}
\input{sections/methods}
\input{sections/experiment}     
\input{sections/conclusion}

\bibliographystyle{ACM-Reference-Format}
\bibliography{sample-base}

\newpage
\appendix
\input{sections/appendix}

\end{document}

%% file: sections/abstract.tex
\begin{abstract}
Proteins, as essential biomolecules, play a central role in biological processes, including metabolic reactions and DNA replication. Accurate prediction of their properties and functions is crucial in biological applications. Recent development of protein language models (pLMs) with supervised fine tuning provides a promising solution to this problem. However, the fine-tuned model is tailored for particular downstream prediction task, and achieving general-purpose protein understanding remains a challenge. In this paper, we introduce Structure-Enhanced Protein Instruction Tuning (SEPIT) framework to bridge this gap. Our approach incorporates a novel structure-aware module into pLMs to enrich their structural knowledge, and subsequently integrates these enhanced pLMs with large language models (LLMs) to advance protein understanding. In this framework, we propose a novel instruction tuning pipeline. First, we warm up the enhanced pLMs using contrastive learning and structure denoising. Then, caption-based instructions are used to establish a basic understanding of proteins. Finally, we refine this understanding by employing a mixture of experts (MoEs) to capture more complex properties and functional information with the same number of activated parameters. Moreover, we construct the largest and most comprehensive protein instruction dataset to date, which allows us to train and evaluate the general-purpose protein understanding model. Extensive experiments on both open-ended generation and closed-set answer tasks demonstrate the superior performance of SEPIT over both closed-source general LLMs and open-source LLMs trained with protein knowledge. 
\end{abstract}

%% file: sections/introduction.tex
\vspace{-4pt}
\section{Introduction}
Proteins are large biomolecules and macromolecules composed of one or more long chains of amino acid residues. They play pivotal roles in catalyzing metabolic reactions, facilitating DNA replication, and driving other essential biological processes~\citep{protein,protein_define}. Generally, proteins are represented in two types of forms: a \textit{one-dimensional (1D) sequence}, which specifies the order of amino acids, and a \textit{three-dimensional (3D) structure} which depicts the spatial configuration of the protein. A protein's 1D sequence is generated by transcribing and translating a gene's DNA sequence and subsequently folds into a specific 3D structure. This 3D conformation determines the protein's properties and functions~\citep{Biomolecule_multimodal}. Conventional machine learning methods~\citep{tradition1,tradition2} have achieved notable accuracy in protein property and function prediction via supervised learning. However, these methods are all task-specific as each model is restricted to predicting a particular property. The growing demand for comprehensive protein analysis in fields such as pathology and drug discovery~\citep{proteinchat} highlights the need for general-purpose protein understanding models capable of accurately predicting a wide range of protein properties and functions.

In the fast-evolving era of large language models (LLMs)~\citep{llm,language_model,tokenselect}, significant efforts have been made to leverage their semantic understanding and complex reasoning capabilities to achieve general-purpose protein property and function prediction. Initially, some approaches treat the 1D protein sequence as natural language input to LLMs~\citep{instructprotein,BioMedGPT,Galactica,Mol-Instructions,BioT5}. However, they focus on learning associations between protein sequences and their properties or functions using only a limited subset of real-world protein sequences, which hinders the LLMs from generating reliable predictions at an evolutionary scale. To address this limitation, ProtST~\citep{Protst} and ProteinCLAP~\citep{Text-guided} utilize pLMs pre-trained on evolutionary-scale protein databases as protein sequence encoders and further employ contrastive learning~\citep{CLIP} on protein-text paired data to incorporate functional information from text with high-quality protein representations from pLMs. 
Unfortunately, these methods remain restricted to prediction and retrieval tasks. However, real-world proteins exhibit complex and diverse properties and functions that require a more comprehensive understanding in an open-ended generative manner rather than being confined to specific tasks. To bridge this gap, Prot2Text~\citep{Prot2text}, ProteinChat~\citep{proteinchat} and ProtT3~\citep{prott3} propose protein-to-language generation by integrating protein sequence or structure encoding from pre-trained models.  

Despite these advancements, existing methods still face significant challenges in providing reliable general-purpose protein understanding for scientific research applications. Firstly, although some studies acknowledge the critical role of a protein's 3D structure in determining its properties and functions, proteins with directly available 3D structural information are rare. As a result, models must learn the relationship between 1D sequences and functional information while relying on limited 3D data, making it difficult to generate accurate property and function predictions. Secondly, existing protein-related instruction datasets overlook 3D structural information and provide limited coverage of protein properties and functions. This gap hinders the comprehensive evaluation of model reliability and generalizability in general-purpose protein understanding tasks. Lastly, the vast diversity of protein properties and functions presents a major challenge for accurate prediction. Using a single general-purpose model to predict a wide range of complex properties and functions is more challenging than fine-tuning multiple specialized models for specific tasks.

To address these challenges, we propose a novel instruction tuning framework called Structure-Enhanced Protein Instruction Tuning (SEPIT) for general-purpose protein understanding. In this framework, we first design a structure-aware module within pLMs to obtain a protein sequence/structure-fused encoder, enabling support for different types of protein inputs (1D or 1D\&3D). Following this, we implement a three-stage training pipeline. In stage 0 (warm-up stage), we train the protein sequence/structure-fused encoder using protein-text contrastive learning and structure denoising, allowing it to leverage a limited amount of structural information to enhance the understanding of large-scale sequence-only proteins. Then, we perform protein instruction tuning to equip LLMs for general-purpose protein understanding capability: in stage 1, we instill fundamental understanding of proteins into the model through protein caption instructions; in stage 2, we initialize mixture of experts (MoEs) through upcycling~\citep{Upcycling}, which allows the model to learn more complex and diverse protein properties and functions while building upon the foundational knowledge from stage 1, all without increasing the number of activated parameters. 
To comprehensively evaluate the reliability and generalizability of our proposed framework, we construct the largest protein instruction dataset to date, which covers the most diverse range of protein properties and functions, based on large-scale protein knowledge bases~\citep{swiss,AlphafoldDB}. In summary, our key contributions are as follows: \begin{itemize}[itemsep=2pt]
\item We designed and integrated a structure-aware module into protein language models, enabling the models to process various types of protein inputs. This significantly improved embedding quality compared to vanilla sequence-only pLMs.
\item We constructed the largest and most extensive protein instruction dataset to date, addressing the gap in datasets for general-purpose protein understanding.
\item We developed a novel protein instruction tuning strategy that allows a single model to learn a broad and complex range of protein properties and functions. This was achieved by leveraging MoEs built upon foundational knowledge.
\item Using the proposed SEPIT framework and our curated protein instruction dataset, we demonstrated the feasibility of equipping large language models with general-purpose protein understanding capability.
\end{itemize}

%% file: sections/relatedwork.tex
\section{Related Work}
In this section, we present the key related work pertinent to our study, focusing primarily on protein language models. Additional related topics including multimodal instruction tuning and learning with 3D structural information are discussed in Appendix \ref{sec:more_related}.

Using context‐aware language models~\citep{language_model}, protein sequences can be treated like sentences, with amino acids acting as individual words. Through pre-training on databases containing hundreds of millions of such protein sequences (\textit{e.g.}, UniRef~\citep{Uniref}, BFD~\citep{BFD1,BFD2}), pLMs enable effective modeling and prediction of protein structures and functions~\citep{pLM_survey}. In earlier works~\citep{unirep,UDSMProt,SeqVec}, LSTM and its variants~\citep{LSTM,LSTM1,LSTM2,LSTM3} were utilized to model the dependencies between residues in single protein sequences. With the rise of the Transformers architecture~\citep{attention}, Transformers-based pLMs emerged. ESM-1b~\citep{ESM1b}, leveraging the Transformers architecture along with a masking strategy for pretraining, significantly enhances the prediction accuracy for mutational effects, secondary structure, and long-range contacts. After this, ProtTrans~\citep{ProtTrans} released two auto-regressive models~\citep{Transformer-XL,XLNet} and four auto-encoder models~\citep{BERT,Albert,Electra,T5} pre-trained on protein sequence databases. Beyond merely focusing on single protein sequences, MSA Transformer~\citep{MSA-Transformer} integrates multiple sequence alignments (MSA) of homologous proteins, providing a solid foundation for the success of AlphaFold2~\citep{AlphaFold2} and AlphaFold3~\citep{AlphaFold3}. Moreover, ESM-2~\citep{ESM2} further scaled up pLMs, achieving protein structure prediction performance comparable to previous works without utilizing MSA information, and significantly reduced inference overhead~\citep{ESM-Fold}. Additionally, there were other studies that attempted to incorporate additional knowledge into the pre-training of protein sequences. For example, ProteinBERT~\citep{ProteinBERT} and OntoProtein~\citep{OntoProtein} incorporate Gene Ontology (GO) annotations directly into their protein embeddings. More recent methods—such as SaProt~\citep{saprot} and ESM-3~\citep{esm3}—go further by fusing both sequence- and structure-based representations, thereby deepening the models’ grasp of protein characteristics. Although these protein language models produce rich, high-quality embeddings, they remain incapable of generating natural-language predictions about protein properties and functions.

%% file: sections/dataset.tex
\section{Construction of Protein Instruction Dataset}

To endow LLMs with general-purpose protein understanding capabilities and evaluate their reliability and generalizability, in this paper, we construct a protein instruction dataset contains open-ended generation and closed-set answer tasks. For the open-ended generation subset, we mainly constructed it based on Swiss-Prot~\citep{swiss}. We include almost all protein properties and functions contained therein (Function, Similarity, Subcellular location, Induction, Molecular Function, Biological Process, Cellular Component, Developmental Stage, Short Sequence Motif, Tissue Specificity, Activity Regulation, Pathway), and utilized ChatGPT~\citep{GPT4} to aid in designing question templates based on the structured annotations. For the closed-set answer subset, we constructed it mainly based on the RCSB PDB~\citep{rcsbpdb}. We follow the data organized by previous researchers~\citep{proteinchat} and select parts of their proposed Q\&A samples that are highly related to protein properties and functions, filtering out other samples related to metadata (\textit{e.g.} discovery time and discovery methods). We have also sampled some examples related to Enzyme Commission (EC) and Gene Ontology (GO) predictions~\citep{ECGO} for inclusion in the closed-set answer subset. More detailed information about the dataset, including statistical information and examples, is shown in Appendix \ref{sec:data}.

Compared to previously proposed protein-text datasets~\citep{Protst,Mol-Instructions,instructprotein,proteinchat}, our dataset offers several key advantages: First, our dataset contains the most comprehensive set of instructions, covering nearly all critical protein properties and function types found in databases~\citep{swiss}. Second, with over 10 million instructions, along with an additional 5 million supplementary instructions from TrEMBL, our dataset is the largest of its kind. Third, by incorporating structural data, our dataset enhances prediction reliability and provides a more robust foundation for protein understanding. In summary, our dataset serves as a valuable resource for advancing research in general-purpose protein understanding.

%% file: sections/methods.tex
\begin{figure*}[t]
  \centering
  \includegraphics[width=1\textwidth]{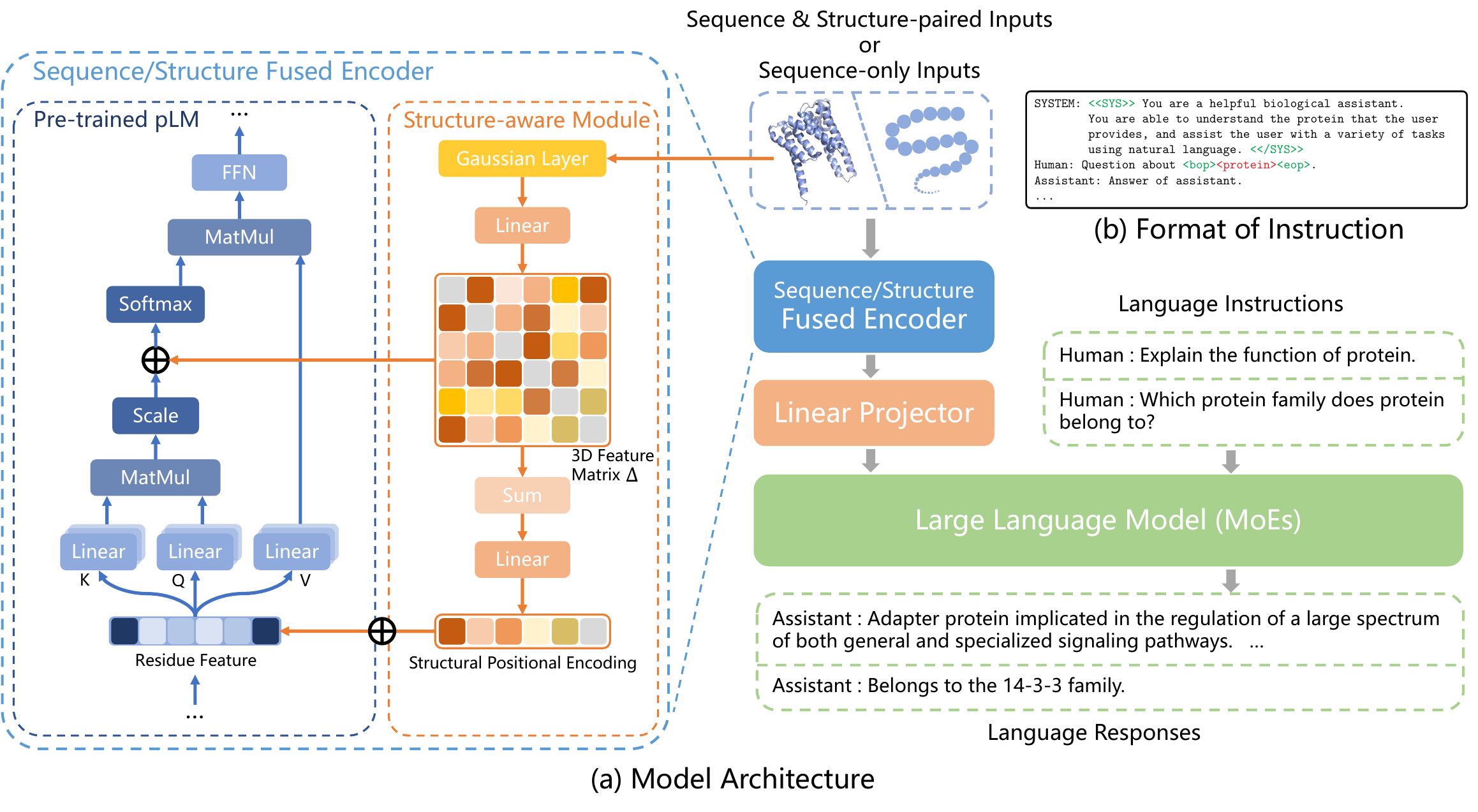}
  \vspace{-20pt}
  \caption{(a) The model architecture of the SEPIT framework includes sequence/structure fused protein encoder, linear projector, and LLMs with MoEs modules, (b) example of instruction format.}
  \label{fig:arch}
  \vspace{-4pt}
\end{figure*}

\section{Structure-Enhanced Instruction Tuning}
\label{sec:method}
In this section, we provide a detailed introduction to our proposed SEPIT framework from two perspectives: model architecture and training pipeline.

\subsection{Model Architecture}
In this subsection, we introduce three main components of the SEPIT framework: 1) a sequence/structure fused protein encoder, 2) a linear projector, and 3) a large language model with MoEs modules. The model architecture is shown in Figure \ref{fig:arch}.

\paragraph{\textbf{Sequence/Structure Fused Protein Encoder.}}
Given the vast availability of sequence-only data and limited sequence-structure paired data (from both experimental and computational sources) within protein domain~\citep{AlphafoldDB,BFD1,BFD2,Uniref}, we propose a sequence/ structure fused protein encoder that is capable of accommodating inputs in either form. Additionally, through leveraging the scarce sequence-structure paired data, we aim to enhance the model's performance when dealing with sequence-only inputs. 

For sequence-only data, numerous pLMs~\citep{ESM1b,ESM2,ProteinBERT,ProtTrans,MSA-Transformer} have already been pre-trained on it. To facilitate their perception of structural information, we have designed a structure-aware module for pLMs. This module encodes the 3D structural information into the attention matrix and positional encoding of pLMs in a simple but effective way. 
Specifically, the attention mechanism of pLMs is defined as follows (for simplicity, we discuss the scenario with single-head and assume that the dimensions of the query, key, and value are all equal to the hidden size $d$). Let $\boldsymbol{X}^{(l)} = [\boldsymbol{x}_1^{(l)}, \boldsymbol{x}_2^{(l)}, \cdots , \boldsymbol{x}_N^{(l)}]^\top$ denote the input to self-attention module in $l$-th layer, where $\boldsymbol{x}_i^{(l)} \in \mathbb{R}^{d}$ is the $d$-dimension representation of the $i$-th residue out of the $N$ residues in the protein. The self-attention module then works as follows:
\begin{small} 
\begin{align}
    \boldsymbol{A}^{(l)}=&\frac{\boldsymbol{X}^{(l)}\boldsymbol{W}_Q^{(l)} (\boldsymbol{X}^{(l)}\boldsymbol{W}^{(l)}_K)^{\top}}{\sqrt{d}}, \\
    \operatorname{Attn}(\boldsymbol{X}^{(l)})=&\operatorname{softmax}(\boldsymbol{A}^{(l)})\boldsymbol{X}^{(l)}\boldsymbol{W}^{(l)}_V,
\label{eq:attention}
\end{align}
\end{small}where $\boldsymbol{W}^{(l)}_Q\in\mathbb{R}^{d\times d},\boldsymbol{W}^{(l)}_K\in\mathbb{R}^{d\times d}, \boldsymbol{W}^{(l)}_V\in\mathbb{R}^{d\times d}$, $\boldsymbol{A}^{(l)}$ is the attention matrix, $\boldsymbol{A}_{i,j}^{(l)}$ denotes the similarity between residue $i$ and $j$. Inspired by previous work on geometric Transformers~\citep{Uni-mol,Transformer-M}, our structure-aware module takes the 3D coordinates $\boldsymbol{C} = [\boldsymbol{c}_1, \boldsymbol{c}_2, \cdots , \boldsymbol{c}_N]^\top, \;\boldsymbol{c} \in \mathbb{R}^3$ of residues (alpha carbon atoms) as input and outputs the 3D feature matrix $\boldsymbol{\Delta} \in \mathbb{R}^{N\times N}$, representing the pairwise spatial relationships of residues in 3D space:
\begin{small} 
\begin{equation}
    \boldsymbol{\Delta} = \phi\left(\boldsymbol{\psi}_{(i,j)}\boldsymbol{W}_a\right)\boldsymbol{W}_b,
    \label{eq:Delta}
\end{equation}
\end{small}where $\boldsymbol{W}_a\in\mathbb{R}^{K\times K},\boldsymbol{W}_b\in\mathbb{R}^{K\times1}$ are linear transformation and $\boldsymbol{\psi}_{(i,j)}=[\psi_{(i,j)}^1,\cdots,\psi_{(i,j)}^K]^\top$ is the Euclidean distance for each twosome of residues undergoes a transformation via the Gaussian Basis Kernel function~\citep{GK}:
\begin{small} 
\begin{equation}
    \psi_{(i,j)}^k = -\frac1{\sqrt{2\pi}|\sigma^k|}\exp\left(-\frac12\left(\frac{\|\boldsymbol{c}_i-\boldsymbol{c}_j\|-\mu^k}{|\sigma^k|}\right)^2\right),\;k=1,...,K,
    \label{eq:GBKF}
\end{equation}
\end{small}where the learnable parameters $\mu^k$ and $\sigma^k$ correspond to the center and scaling coefficient of the $k$-th Gaussian Basis Kernel. These relationships are incorporated into the attention matrix as bias: 
\begin{small}
\begin{equation}
    \hat{\boldsymbol{A}}^{(l)} = \boldsymbol{A}^{(l)} + \boldsymbol{\Delta},
    \label{eq:attn_bias}
\end{equation}
\end{small}and added as structure positional encoding to the input of pLMs:
\begin{small}
\begin{equation}
    \hat{\boldsymbol{X}}^{(0)} = \boldsymbol{X}^{(0)} + \omega\left(\sum_{j\in[n]}\boldsymbol{\psi}_{(i,j)}\right)\boldsymbol{W}_c,
    \label{eq:input_of_pLM}
\end{equation}
\end{small}where $\omega$ signifies the coefficient that regulates the magnitude of the structure positional encoding and $\boldsymbol{W}_c\in\mathbb{R}^{K\times d}$ is learnable linear transformation. It is noteworthy that when the input to the sequence/structure fused protein encoder consists solely of the protein sequence, lacking structural information, the structure-aware module will be automatically disabled. This allows it to adapt to different protein input formats.

\begin{figure*}[t]
  \centering
  \includegraphics[width=0.98\textwidth]{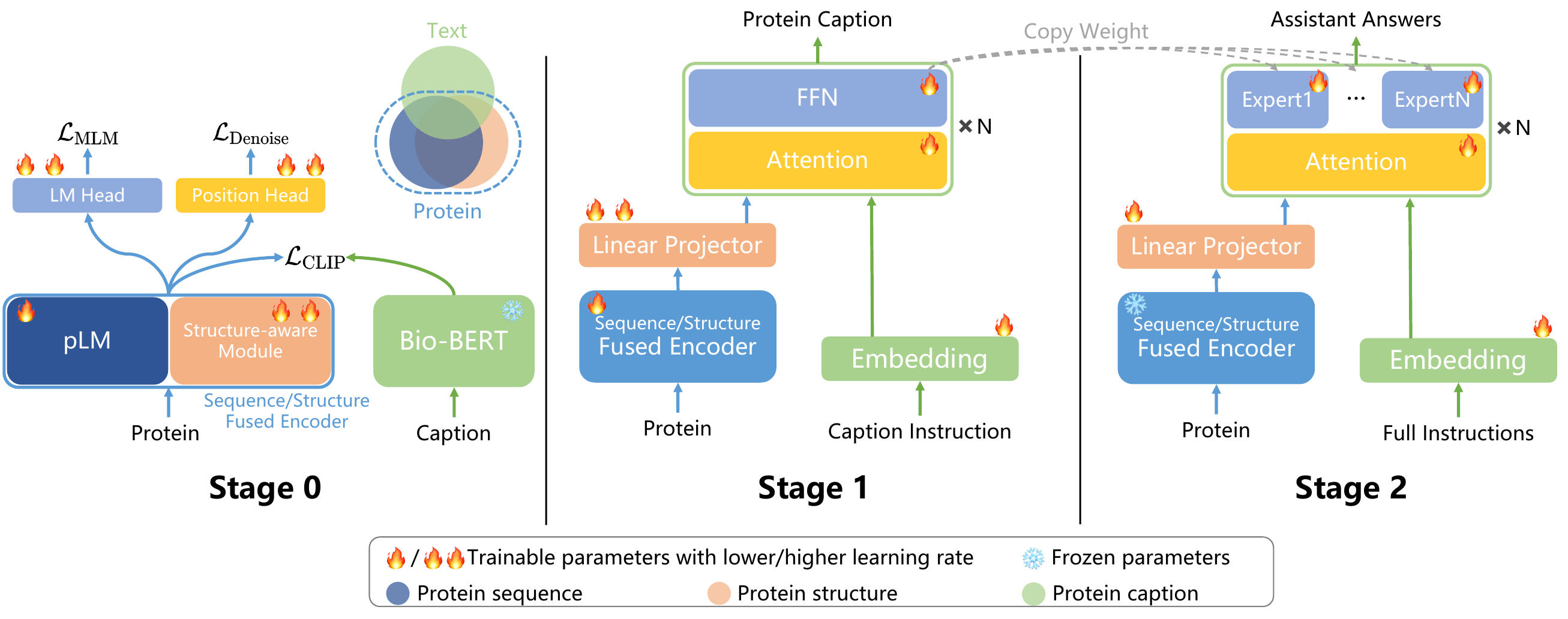}
   \vspace{-8pt}
  \caption{The three-stage training pipeline of SEPIT with a warm-up stage (stage 0) for protein encoder and two instruction tuning stages (stage 1 \& stage 2).}
   \vspace{-8pt}
  \label{fig:pipe}
\end{figure*}

\paragraph{\textbf{Linear Projector.}}
To bridge proteins with natural language, a module is required to link the protein encoder and the LLMs decoder. Prior work in the MLLMs field has contributed outstanding methods such as Q-former~\citep{Blip-2}, linear projector~\citep{LLaVA}, and merging tokens before the linear projector~\citep{Minigpt-4}. Considering the vast differences between proteins and visual images - that is, the former requires the retention of more information of all residues (as any change in the amino acid sequence can lead to significant structural differences, resulting in profoundly different properties and functions), whereas the latter possesses some degree of information redundancy - we opt for a simple linear projector to reduce information loss.

\paragraph{\textbf{Large Language Model with Mixture of Experts Module.}}
Due to the understanding of proteins being a complex multi-task problem, the various properties and functions of proteins can exhibit significant changes with subtle variations in the amino acid sequence. As the hub of "understanding" within the entire framework, the capabilities of LLMs are crucial. Previous scaling laws~\citep{Scaling_law} have suggested that larger parameter sizes can endow LLMs with stronger capabilities; however, the additional computation costs brought about by increased activated parameters are intolerable for us. Therefore, we seek to leverage mixture-of-experts (MoEs) to achieve higher parameter sizes without increasing the number of activated parameters, thereby enhancing the model's capacity and generalization ability. In our framework, the MoEs module replaces the FFN module in each Transformer decoder layer. The MoEs module works as follows~\citep{Gshard,adaptive,ST-MoE,MoE-LLaVa,moe-rec,ReFound}:
\begin{small}
\begin{align}
y = \sum_{i=1}^{n_e}\text{G}(\boldsymbol{x})_i\cdot\text{E}_i(\boldsymbol{x}).
\end{align}
\end{small}Here, $\boldsymbol{x}$ is assumed to be the input to the original FFN layer, and $\text{E}$ represents the $n_e$ experts in the MoEs, each of which has the exact same structure as the original FFN layer. $\text{G}$ represents the gating network, $\text{G}(\boldsymbol{x})_i$ denotes the gate weight for the $i$-th expert, and $\text{E}_i(\boldsymbol{x})$ is the output of the $i$-th expert. For the gating network, we employ the commonly used linear TopK gate:
\begin{small}
\begin{equation}
    \text{G}(\boldsymbol{x}):=\operatorname{Softmax}\left(\operatorname{TopK}\left(\boldsymbol{x} \cdot \boldsymbol{W}_{\text{G}}\right)\right),
\end{equation}
\end{small}and we impose auxiliary loss to ensure the token balance among the experts~\citep{ST-MoE,Gshard}:
\begin{small}
\begin{equation}
    \mathcal{L}_\text{aux} = n \cdot \sum_{i=1}^{n} f_{i} \cdot P_{i},
    \label{eq:aux}
\end{equation}
\end{small}where $f_i$ is the proportion of tokens processed by expert $i$ and $p_i$ represents the proportion of gating weight allocated to an expert. 

\subsection{Training Pipeline}

In this subsection, we will discuss the details of the training pipeline for SEPIT framework, based on the model architecture presented before. As shown in Figure \ref{fig:pipe}, the whole pipeline includes three stages: in stage 0, we warm up our proposed sequence/structure fused protein encoder based on pre-trained pLM primarily through protein-text contrastive learning and structure denoising. In stage 1, we pre-train on the protein captioning task to further align protein representations with natural language, while concurrently infusing foundational protein knowledge into LLMs. In stage 2, we initiate the MoEs modules using the sparse upcycling~\citep{Upcycling} from the FFNs of the LLMs trained in stage 1 and perform instruction tuning on our proposed protein instruction dataset.

\paragraph{\textbf{Stage 0: Warming Up the Sequence/Structure Fused Protein Encoder.}}
In this stage, our primary goal is to warm up our protein encoder. Although the pLM is already pre-trained, the structure-aware module that we plug in is randomly initialized. To address this, we leverage two main pre-training paradigms. Firstly, to enable the structure-aware module to better perceive structural information, we follow the common practice in molecular self-supervised learning~\citep{Mol_pretrain1,Mol_pretrain2,Transformer-M,Uni-mol}, which involves structure denoising tasks. For the input 3D coordinates of protein residues $\boldsymbol{C} = [\boldsymbol{c}_1, \boldsymbol{c}_2, \cdots , \boldsymbol{c}_N]^\top$, we apply noise to obtain noised coordinates $\tilde{\boldsymbol{C}} = [\tilde{\boldsymbol{c}}_1, \tilde{\boldsymbol{c}}_2, \cdots , \tilde{\boldsymbol{c}}_N]^\top$, where $\tilde{\boldsymbol{c}}_i = \boldsymbol{c}_i + \alpha\boldsymbol{\delta}_i,\; \boldsymbol{\delta}_i \sim\mathcal{N}(0,I)$, $\alpha$ is a scaler used to control the magnitude of noise. Then we predict the applied noise based on them (equivalent to predicting the original 3D coordinates). The formula for denoise loss is as follows:
\begin{small}
\begin{equation}
    \mathcal{L}_\text{Denoise}=\frac{1}{3n}\sum_{i=1}^n\sum_{j=1}^3\left(\boldsymbol{\delta}_i^j-\hat{\boldsymbol{\delta}_i^j}\right)^2,
\end{equation}
\end{small}where $\hat{\boldsymbol{\delta}_i^j}$ is output by an additional SE(3) equivariant attention layer~\citep{Transformer-M,Denoise} (position head), which takes the last hidden state of the protein encoder $\boldsymbol{X}^{(L+1)}$ and $\boldsymbol{\Delta}$ in Equation~\ref{eq:attn_bias} as input:
\begin{small}
\begin{equation}
    \hat{\boldsymbol{\delta}}_i^j=\left(\sum_{i=1}^n\boldsymbol{\Delta}_{ik}\boldsymbol{D}_{ik}^j\boldsymbol{X}^{(L+1)}_k\boldsymbol{W}_m\right)\boldsymbol{W}_n,\quad \boldsymbol{D}_{ik}=\frac{\mathbf{c}_{i}-\mathbf{c}_{k}}{\|\mathbf{c}_{i}-\mathbf{c}_{k}\|}.
\end{equation}
\end{small}Secondly, we utilize protein-text contrastive learning to further promote the fusion of protein sequence and structure information under text supervision, while concurrently aligning protein representations with their textual descriptions. Formally, given a batch of paired proteins and protein captions $\{(\mathcal{P}_i, \mathcal{T}_i)\}_{i=1}^B$, the CLIP loss can be expressed as:
\begin{small}
\begin{equation}
    \mathcal{L}_{\text{CLIP}}=-\frac1{2B}\sum_{i=1}^B\left(\log\frac{\exp(\boldsymbol{p}_i\cdot \boldsymbol{t}_i/\tau)}{\sum_{j=1}^B\exp(\boldsymbol{p}_i\cdot \boldsymbol{t}_j/\tau)}+\log\frac{\exp(\boldsymbol{p}_i\cdot \boldsymbol{t}_i/\tau)}{\sum_{j=1}^B\exp(\boldsymbol{p}_j\cdot \boldsymbol{t}_i/\tau)}\right),
\end{equation}
\end{small}where $\boldsymbol{p}_i$,$\boldsymbol{t}_i$ are the representations of $P_i$ and $T_i$ output by the protein encoder and text encoder (BERT), respectively. Additionally, we also maintain the Masked Language Model (MLM) training objective related to the protein sequence (consistent with ESM2~\citep{ESM2}) as a regularization term to prevent catastrophic forgetting in the pLM. Overall, the loss for stage 0 is as follows:
\begin{small}
\begin{equation}
    \mathcal{L}_\text{stage 0} = \mathcal{L}_\text{Denoise}+\mathcal{L}_{\text{CLIP}}+\mathcal{L}_\text{MLM}.
\end{equation}
\end{small}Under the influence of these training objectives, the mutual information between protein sequence and protein structure as well as protein and text is increased.

\input{tables/table1}
\paragraph{\textbf{Stage 1: Pre-training on Protein Captions.}}
In this stage, our fundamental objective is to further align proteins with their natural language descriptions through the paradigm of conditional generation~\citep{LLaVA,instructGPT}, utilizing protein caption instructions. To ensure consistency in the model's handling of different forms of protein inputs, we randomly input protein data, both those with only sequences and those paired with structures, into our protein encoder at probabilities of $15\%$ and $85\%$, respectively. The output protein representation sequences are mapped into the textual space through linear projector, in conjunction with protein caption instructions to guide LLMs in producing straightforward descriptions of proteins, such as function, family, subcellular localization, and overall descriptions. Formally, consider a protein-text pair $(\mathcal{P}, \mathcal{T})$ similar to that in stage 0, given the output sequence of the protein encoder $\boldsymbol{S_p}$, and instructions $\boldsymbol{S_\text{instruct}}$, the objective of stage 1 is as follows:
\begin{small}
\begin{equation}
    \mathcal{L}_\text{stage 1} = p(\boldsymbol{S_t}|f(\mathbf{S_p}),\boldsymbol{S_\text{instruct}})=\prod_{i=1}^Lp_{\boldsymbol{\theta}}(s_i|\boldsymbol{S_p},\boldsymbol{S}_{\text{instruct},<i},\boldsymbol{S}_{\boldsymbol{t},<i}),
\end{equation}
\end{small}where $f(\cdot)$ denotes the linear projector, $\theta$ represents all trainable parameters, $\boldsymbol{S}_{\text{instruct},<i}$ and $\boldsymbol{S}_{\boldsymbol{t},<i}$ respectively signify the instruction and answer tokens preceding the current prediction token $s_i$, and $L$ denotes the total sequence length accepted by LLMs.

\paragraph{\textbf{Stage 2: Upcycling and Instruction Tuning.}}
In this stage, our main aim is to upcycle the model obtained from stage 1 by replacing each FFN module within the LLMs with MoE module, where each expert is initialized by an FFN. In the case of top$-1$ activation, this approach offers a larger model parameter count under the same activation parameter volume. Meanwhile, the basic understanding of proteins acquired in stage 1 lays the groundwork for more complex and multifaceted learning in this stage. Based on this, we utilize a diverse set of protein instructions for instruction tuning, aiming to endow SEPIT with general-purpose protein understanding capabilities. Similar to stage 1, the loss function for stage 2 can be formally represented as:
\begin{small}
\begin{equation}
    \mathcal{L}_\text{stage 2} = \mathcal{L}_\text{stage 1} + \beta \mathcal{L}_\text{aux},
\end{equation}
\end{small}where $\mathcal{L}_\text{aux}$ is an auxiliary loss used for constraining the token balance among experts, as mentioned in Equation~\ref{eq:aux}, with $\beta$ being used to control its relative magnitude.

%% file: tables/table1.tex
\begin{table*}[t]
  \small
  \centering
  \renewcommand{\arraystretch}{1.1}
  \caption{Performance comparisons on open-ended generation and closed-set answer tasks.}
  \vspace{-4pt}
  \begin{adjustbox}{width=1\textwidth}
  \begin{tabular}{lccccccccccc}
    \toprule
    \multicolumn{1}{l}{\multirow{2}{*}{\textbf{Model}}} &
      \multicolumn{1}{c}{\multirow{2}{*}{\textbf{\begin{tabular}[c]{@{}c@{}}Activated \\ Parameters\end{tabular}}}} &
      \multicolumn{9}{c}{\textbf{Open-Ended}} &
      \textbf{Closed-Set} \\ \cmidrule(lr){3-11} \cmidrule(lr){12-12} 
    \multicolumn{1}{l}{} &
      \multicolumn{1}{c}{} &
      \textbf{BLEU-2} &
      \textbf{BLEU-4} &
      \textbf{ROUGE-1} &
      \textbf{ROUGE-2} &
      \textbf{ROUGE-L} &
      \textbf{METEOR} &
      \textbf{BERT-P} &
      \textbf{BERT-R} &
      \multicolumn{1}{c}{\textbf{BERT-F1}} &
      \textbf{Accuracy} \\ \midrule
    \multicolumn{12}{c}{\textbf{Zero-Shot}} \\ \midrule
    \multicolumn{1}{l}{GPT-3.5-turbo} &
      \multicolumn{1}{c}{N/A} &
      3.26 &
      0.02 &
      12.41 &
      3.14 &
      11.06 &
      10.44 &
      85.18 &
      85.40 &
      \multicolumn{1}{c}{85.24} &
      56.56\% \\
    \multicolumn{1}{l}{Claude-3-haiku} &
      \multicolumn{1}{c}{N/A} &
      3.00 &
      0.07 &
      12.10 &
      2.65 &
      10.62 &
      9.28 &
      86.04 &
      85.47 &
      \multicolumn{1}{c}{85.70} &
      59.14\% \\
    \multicolumn{1}{l}{GPT-4-turbo} &
      \multicolumn{1}{c}{N/A} &
      4.21 &
      0.08 &
      12.78 &
      2.93 &
      11.57 &
      10.41 &
      86.91 &
      85.56 &
      \multicolumn{1}{c}{85.71} &
      58.58\% \\
    \multicolumn{1}{l}{GPT-4o-mini} &
      \multicolumn{1}{c}{N/A} &
      2.36 & 0.01 & 10.72 & 1.93 & 9.03 & 9.00 & 85.08 & 85.38 & \multicolumn{1}{c}{85.19} & 41.58\% \\
    \multicolumn{1}{l}{GPT-4o} &
      \multicolumn{1}{c}{N/A} &
      3.26 & 0.14 & 15.02 & 3.73 & 12.30 & 13.01 & 85.28 & 86.51 & \multicolumn{1}{c}{85.86} & 59.70\% \\
    \multicolumn{1}{l}{DeepSeek-V3} &
      \multicolumn{1}{c}{671B} &
      2.75 & 0.02 & 12.15 & 3.37 & 10.54 & 10.25 & 85.32 & 85.63 & \multicolumn{1}{c}{85.44} & 56.49\% \\
    \multicolumn{1}{l}{Galactica} &
      \multicolumn{1}{c}{1.3B} &
      0.43 &
      0.01 &
      3.49 &
      0.41 &
      2.67 &
      2.44 &
      85.79 &
      82.61 &
      \multicolumn{1}{c}{84.08} &
      39.15\% \\
    \multicolumn{1}{l}{BioMedGPT} &
      \multicolumn{1}{c}{7B} &
      0.83 &
      0.01 &
      4.90 &
      0.49 &
      3.26 &
      4.59 &
      85.51 &
      84.95 &
      \multicolumn{1}{c}{85.14} &
      38.61\% \\ 
    \multicolumn{1}{l}{Mol-Instructions} &
      \multicolumn{1}{c}{7B} &
      0.53 &
      0.01 &
      5.96 &
      0.39 &
      4.64 &
      5.51 &
      83.81 &
      84.41 &
      \multicolumn{1}{c}{84.06} &
      --- \\ 
      \multicolumn{1}{l}{BioT5+} &
      \multicolumn{1}{c}{252M} &
      3.88 &
      1.92 &
      12.12 &
      4.88 &
      10.37 &
      14.26 &
      85.14 &
      85.93 &
      \multicolumn{1}{c}{85.48} &
      --- \\
      \multicolumn{1}{l}{InstructProtein} &
      \multicolumn{1}{c}{1.3B} &
      5.50 &
      2.97 &
      14.80 &
      5.68 &
      13.76 &
      13.17 &
      85.34 &
      85.92 &
      \multicolumn{1}{c}{85.57} &
      48.37\% \\ \midrule
    \multicolumn{12}{c}{\textbf{Instruction Tuning}} \\ \midrule
    \multicolumn{1}{l}{TinyLlama} &
      \multicolumn{1}{c}{1.1B} &
      51.16 &
      43.44 &
      65.41 &
      51.26 &
      62.31 &
      60.80 &
      93.97 &
      94.37 &
      \multicolumn{1}{c}{94.16} &
      74.09\% \\
    \multicolumn{1}{l}{OpenLlama-v2} &
      \multicolumn{1}{c}{3B} &
      36.19 &
      30.65 &
      48.33 &
      36.52 &
      45.53 &
      49.01 &
      92.92 &
      91.87 &
      \multicolumn{1}{c}{92.35} &
      71.77\% \\
    \multicolumn{1}{l}{Llama2} &
      \multicolumn{1}{c}{7B} &
      57.02 &
      49.47 &
      70.80 &
      57.24 &
      67.78 &
      65.96 &
      94.95 &
      95.17 &
      \multicolumn{1}{c}{95.05} &
      71.68\% \\ \midrule
    \multicolumn{12}{c}{\textbf{Sequence-Only Protein Instruction Tuning}} \\ \midrule
    \multicolumn{1}{l}{PIT-TinyLlama} &
      \multicolumn{1}{c}{1.8B} &
      57.82 &
      50.02 &
      71.34 &
      58.16 &
      68.35 &
      66.19 &
      95.18 &
      95.28 &
      \multicolumn{1}{c}{95.26} &
      76.02\% \\
    \multicolumn{1}{l}{PIT-TinyLlama-MoEs} &
      \multicolumn{1}{c}{1.8B} &
      57.92 &
      50.01 &
      72.13 &
      58.21 &
      69.19 &
      66.29 &
      95.31 &
      95.30 &
      \multicolumn{1}{c}{95.29} &
      78.56\% \\ \midrule
    \multicolumn{12}{c}{\textbf{Structure-Enhanced Protein Instruction Tuning}} \\ \midrule
    \multicolumn{1}{l}{SEPIT-TinyLlama} &
      \multicolumn{1}{c}{1.8B} &
      58.43 &
      51.04 &
      72.34 &
      58.77 &
      69.13 &
      67.91 &
      95.32 &
      95.59 &
      \multicolumn{1}{c}{95.44} &
      79.05\% \\
    \multicolumn{1}{l}{\textbf{SEPIT-Llama2}} &
      \multicolumn{1}{c}{\underline{8B}} &
      \textbf{60.81} &
      \textbf{52.37} &
      \textbf{74.80} &
      \textbf{60.84} &
      \textbf{71.62} &
      \textbf{68.43} &
      \textbf{95.81} &
      \textbf{95.73} &
      \multicolumn{1}{c}{\textbf{95.76}} &
      \textbf{79.97\%} \\
    \multicolumn{1}{l}{\textbf{SEPIT-TinyLlama-MoEs}} &
      \multicolumn{1}{c}{\textbf{1.8B}} &
      \underline{60.28} &
      \underline{52.16} &
      \underline{74.22} &
      \underline{60.29} &
      \underline{71.13} &
      \underline{68.27} &
      \underline{95.62} &
      \underline{95.69} &
      \multicolumn{1}{c}{\underline{95.64}} &
      \underline{ 79.73\%} \\ \bottomrule
    \end{tabular}
    \end{adjustbox}
    \vspace{-4pt}
    \label{tab:performance}
\end{table*}

%% file: sections/experiment.tex
\input{tables/table2}
\section{Experiments}
\label{sec:exp}
In this section, we will first outline the experimental setting. Subsequently, we will demonstrate SEPIT's effectiveness and design benefits via performance comparisons and ablation studies. Finally, we will delve deeper into the characteristics of SEPIT through generalization analysis and case studies. Additional analysis, including error pattern analysis are provided in the Appendix \ref{sec:more_ana}.

\subsection{Experimental Setting}

First, we provide a brief overview of the main experimental settings, with more detailed information available in Appendix \ref{sec:data}, \ref{sec:more_imple}.

\paragraph{\textbf{Use of Pre-training Data.}}
In each stage of SEPIT, we utilize our proposed protein instruction dataset, employing different subsets at various stages. At Stage 0, our focus is primarily on basic protein descriptions derived from Swiss-Prot and the RCSB PDB. For Swiss-Prot, akin to previous work~\citep{Protst}, we formulate the protein captions using Function, Similarity and Subcellular location. Regarding the RCSB PDB, we directly utilize abstracts from related PubMed papers collected by ~\citep{proteinchat} as captions. In Stage 1, we employ the same data as in Stage 0, but the output format is altered to the style of caption instructions. During Stage 2, we utilize the complete protein instruction dataset we proposed, which includes open-ended generation and closed-set answers tasks.

\paragraph{\textbf{Baselines and Evaluation Metrics.}}
We evaluate the capability of SEPIT for general-purpose protein understanding on the test set of the protein instruction dataset we proposed with the state-of-the-art models. There are four main categories of methods. Among the Zero-Shot methods, we include current mainstream LLMs providing API services (e.g., Claude-3-haiku~\citep{Claude-3-haiku}, GPT-3.5-turbo/4-turbo/4o-mini/4o~\citep{GPT4} and DeepSeek-V3~\citep{DeepSeek-V3}), open-source LLMs fine-tuned on biomedical corpus (e.g., Galactica~\citep{Galactica}, BioMedGPT~\citep{BioMedGPT}) and open-source LLMs fine-tuned specifically on molecular or protein knowledge(e.g., Mol-Instructions~\citep{Mol-Instructions}, BioT5+~\citep{BioT5}, InstructProtein~\citep{instructprotein} and ProtT3~\citep{prott3}). In the category of instruction tuning methods, we evaluate mainstream open-source LLMs (e.g., TinyLlama-Chat~\citep{TinyLlama}, OpenLlama-v2~\citep{openllmai23}, Llama2-Chat~\citep{Llama2}), where the protein sequences are input in natural language form. For sequence-only protein instruction tuning methods (PITs), ESM2-650M~\citep{ESM2} was utilized as the protein encoder, with only protein sequences input. In addition to our proposed SEPIT framework, SEPIT-TinyLlama-MoEs, we have also designed two variants that differ in the LLMs' architecture. For evaluation metrics, we employ BLEU score~\citep{BLEU}, ROUGE score~\citep{ROUGE}, METEOR score~\citep{METEOR}, BERT score~\citep{Bertscore} calculated by PubMedBERT~\citep{PubMedBert}, and Accuracy to assess performance across two types of tasks: open-ended generation and closed-set answer, respectively.

\input{tables/table3_1}
\subsection{Performance Comparisons}
The results of performance comparisons are shown in Table \ref{tab:performance}. We can observe that: 1) Our proposed SEPIT consistently outperforms the baseline models by a significant margin. Specifically, SEPIT-Llama achieves the highest performance across all metrics in both types of tasks. In comparison, SEPIT-TinyLlama-MoEs demonstrates significantly higher parameter efficiency, achieving almost identical results to SEPIT-Llama with just $1/6$ of the LLMs' activated parameters. 2) Zero-Shot methods generally perform poorly, with neither powerful general models like GPT and Claude nor open-source models fine-tuned on biomedical corpus or protein knowledge able to accomplish protein understanding tasks well. Notably, Mol-Instructions and BioT5+ are trained on protein-related instructions. However, limited data diversity causes catastrophic forgetting to instruction following, hindering their ability to provide closed-set answers or accurate open-ended responses. 3) Instruction tuning on pure LLMs endows LLMs with decent protein understanding capabilities and demonstrates certain scaling laws~\citep{Scaling_law} (OpenLlama-v2 demonstrates suboptimal results as it has not been specifically fine-tuned for chat assistant purpose.) However, overall, due to the lack of prior knowledge learned from evolutionary-scale protein databases, they can only achieve limited performance. 4) While utilizing prior knowledge from pLMs can significantly enhance the performance of LLMs of the same scale, the lack of structural awareness in PIT only results in suboptimal outcomes compared to our proposed SEPIT.

\input{tables/table3_2}
Considering that there are additional related works~\citep{prollama,proteinchat,prott3} capable of captioning input protein sequences but lacking instruction-following capabilities, we attempted to compare performance on overlapping properties and functions with these works. Table \ref{tab:more_performance} shows a performance comparison with ProtT3~\citep{prott3}, which is the most representative among these works, demonstrates superior performance. For open-ended generation and closed-set answer tasks, we specifically selected Function, Similarity, Subcellular location, and overlapping Q\&A about proteins from RCSB PDB for comparison. The results demonstrate that our model exhibits significantly better performance.

\subsection{Ablation Studies}
In this section, we will explore the impact of various designs of SEPIT on its performance from two aspects: the model and the data.

\paragraph{\textbf{Model Ablation.}} 
For SEPIT's model architecture, we propose the following variants: without the structure-aware module (w/o Structure), without the mixture of experts module (w/o MoEs), without both the aforementioned modules (w/o Structure \& MoEs), without Stage 0 pre-training (w/o Stage 0), and completely excluding the SEPIT framework (w/o SEPIT).
Table \ref{tab:ablation} shows the results of the ablation study, proving the significant effectiveness of each component within SEPIT's model architecture. The absence of either the structure-aware module or the MoEs Module leads to performance degradation in both open-ended generation tasks and closed-set answer tasks, with further deterioration when both are excluded. Meanwhile, the performance under w/o SEPIT intuitively demonstrates the overall effectiveness of the framework. It is noteworthy that the results for w/o Stage 0 are not available, as under the implementation using automatic mixed precision (AMP) based on FP16, the randomly initialized structure-aware module would bring excessively large gradients causing overflow, even though we employed a warm-up learning rate scheduler. Due to device restrictions, we were unable to use BF16 type; however, this issue was resolved as Stage 0 progressed. In order to supplement the analysis of the effectiveness of Stage 0, we validate the protein encoder trained by Stage 0 on commonly used EC, GO annotation tasks~\citep{ECGO}. The results, as shown in Table \ref{tab:ecgo}, demonstrate the performance comparisons to state-of-the-art methods on $\text{F}_\text{max}$ metric.

\paragraph{\textbf{Data Ablation.}}
Regarding the data, apart from using Swiss-Prot and RCSB PDB for constructing the protein instruction dataset, there exists a substantial amount of protein-text paired data in TrEMBL~\citep{swiss}. Considering that the TrEMBL data is annotated by automated methods and has not been manually screened, we select proteins with more comprehensive descriptions (annotation score $\geq 4$) to construct a supplementary dataset using the same method, with a sample size (5.25M) close to the entire protein instruction dataset (5.47M). Disappointingly, as shown in Table \ref{tab:ablation} (w/ TrEMBL), even after doubling the GPUs cost, what we obtain is a decrease in performance. Our analysis suggests that directly mixing low-quality data into high-quality data introduces noise, and protein understanding tasks require higher quality over quantity for data. More results can be found in Appendix \ref{sec:more_ablation}.

\input{tables/table4_1}
\subsection{Generalization Analysis}
\paragraph{\textbf{Generalization Across Different Protein Input Formats.}}
Towards a general-purpose protein understanding capability, SEPIT supports both sequence-only and sequence-structure paired protein inputs, achieving consistent results as shown in Table \ref{tab:consistency}. The three SEPIT variants all yields very similar effects on both types of protein inputs. Moreover, compared to the corresponding scale PIT model, SEPIT demonstrates a stronger understanding of sequence-only inputs. This implies that through SEPIT, we can utilize a small amount of sequence-structure paired data to enhance the understanding of a large volume of sequence-only protein inputs.

\input{tables/table4_2}
\paragraph{\textbf{Generalization to Out-of-Distribution Proteins.}} 
To further evaluate the generalization ability of SEPIT on highly novel proteins, we constructed an out-of-distribution (OOD) test dataset by filtering sequences with a maximum amino acid similarity of $<40\%$ using the CD-HIT clustering algorithm~\cite{CD-HIT}. As shown in Table \ref{tab:model_comparison_ood}, all methods exhibited performance drops on this challenging setup, with instruction tuning methods being particularly affected. However, SEPIT demonstrated stronger generalization compared to alternative approaches, owing to its integration of both pLMs' prior knowledge and the structural information. Remarkably, SEPIT maintained its performance even without structural input during inference, highlighting its robust generalization capacity.

\input{tables/table5}
\begin{figure*}[t]
  \centering
  \includegraphics[width=1\textwidth]{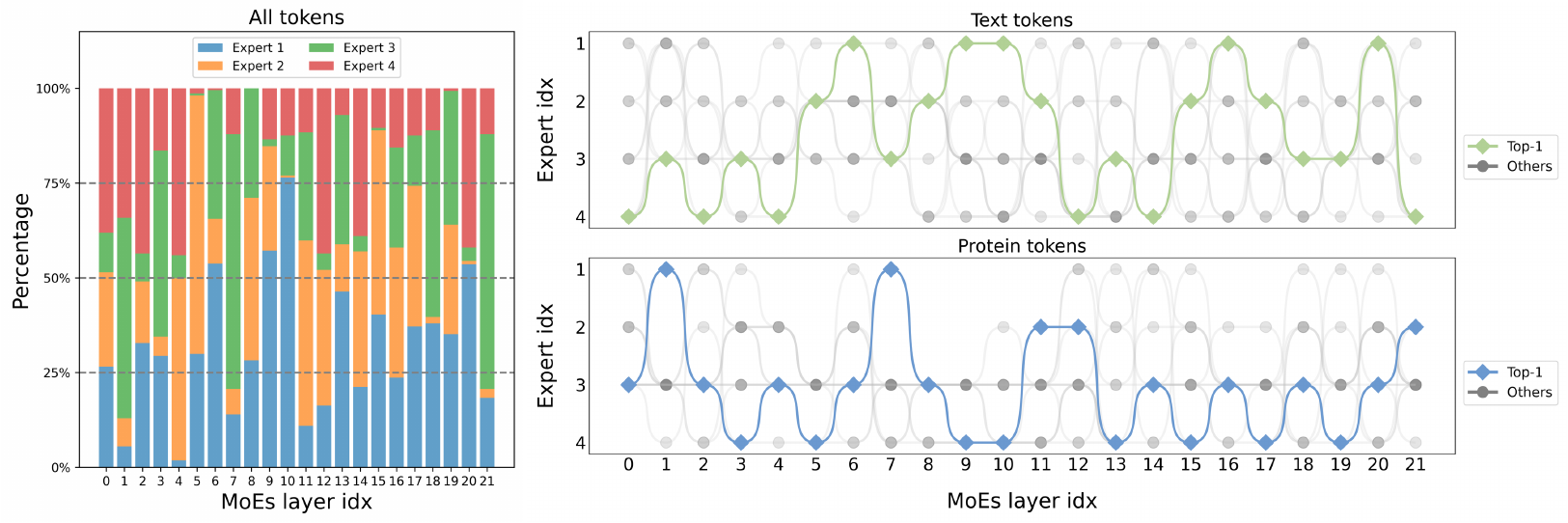}
  \vspace{-20pt}
  \caption{The workload of experts in SEPIT (left) and tokens' pathways among experts (right).}
  \vspace{-8pt}
  \label{fig:moe1}
\end{figure*}

\subsection{Case Studies}
\paragraph{\textbf{General-Purpose Protein Understanding Ability of SEPIT}}
At last, we would like to showcase the general-purpose protein understanding capability of SEPIT. As shown in Table \ref{tab:case}, we present three cases from the test set of our protein instruction dataset. For case 1, SEPIT accurately responds regarding the protein's function, whereas PIT incorporates incorrect details, and both Llama-Chat and GPT-4 offer entirely inaccurate responses. For case 2, SEPIT also gives the correct response, while the answers from PIT and Llama-Chat, although covering the correct answer, come with additional incorrect information, likely due to hallucinations caused by the lack of structural information. For case 3, SEPIT provides the partially correct response, whereas PIT and Llama-Chat exhibit a similar error pattern to case 2. Due to space limitations, more cases and analysis are included in the Appendix \ref{sec:more_case}, \ref{sec:more_ana}.

\paragraph{\textbf{Workload of Experts in SEPIT}}
In SEPIT, we utilize mixture of experts, and in Figure \ref{fig:moe1}, we present the workload distribution of different experts during inference on test set of the open-ended generation task. In the left graph, we can observe that the experts are evenly activated, indicating that the auxiliary loss has played its expected role, which lays the foundation for efficient parallel inference of experts. In the right graph, we visualize the pathways distribution of text and protein tokens across experts in different layers, and we observe an intriguing phenomenon. Unlike the findings in previous vision-language multimodal research~\citep{MoE-LLaVa}, in SEPIT, text and protein tokens are processed by different experts instead of following almost identical pathways as do text and image tokens. We believe that this stems from the fundamental difference between proteins and images. That is, protein tokens, which represent amino acids, cannot reflect the properties of the entire protein, while image tokens represent specific regions of an image containing independent information that can correspond to a part of the image's caption. This validates our choice to use complete protein representation sequences as inputs for LLMs, rather than compressing tokens as is often done in vision-language tasks. More visualization is shown in Appendix \ref{sec:more_case}.

%% file: tables/table2.tex
\begin{table*}[t]
  \small
  \centering
  \caption{Performance comparisons on overlapping properties and functions.}
   \vspace{-4pt}
   \begin{adjustbox}{width=1\textwidth}
  \begin{tabular}{lccccccccccc}
    \toprule
    \multicolumn{1}{l}{\multirow{2}{*}{\textbf{Model}}} &
      \multicolumn{1}{c}{\multirow{2}{*}{\textbf{\begin{tabular}[c]{@{}c@{}}Activated \\ Parameters\end{tabular}}}} &
      \multicolumn{9}{c}{\textbf{Open-Ended}} &
      \textbf{Closed-Set} \\ \cmidrule(lr){3-11} \cmidrule(lr){12-12} 
    \multicolumn{1}{l}{} &
      \multicolumn{1}{c}{} &
      \textbf{BLEU-2} &
      \textbf{BLEU-4} &
      \textbf{ROUGE-1} &
      \textbf{ROUGE-2} &
      \textbf{ROUGE-L} &
      \textbf{METEOR} &
      \textbf{BERT-P} &
      \textbf{BERT-R} &
      \multicolumn{1}{c}{\textbf{BERT-F1}} &
      \textbf{Accuracy} \\ \midrule
    \multicolumn{1}{l}{ProtT3} &
      \multicolumn{1}{c}{1.3B} &
      65.38 &
      51.87 &
      73.75 &
      57.88 &
      72.88 &
      70.39 &
      96.29 &
      95.80 &
      \multicolumn{1}{c}{96.03} &
      90.52\% \\
    \multicolumn{1}{l}{\textbf{SEPIT-TinyLlama-MoEs}} &
      \multicolumn{1}{c}{1.8B} &
      \textbf{83.07} &
      \textbf{81.29} &
      \textbf{86.74} &
      \textbf{83.55} &
      \textbf{86.05} &
      \textbf{85.96} &
      \textbf{98.04} &
      \textbf{98.01} &
      \multicolumn{1}{c}{\textbf{98.02}} &
      \textbf{94.92\%} \\ \bottomrule
    \end{tabular}
    \end{adjustbox}
     \vspace{-4pt}
    \label{tab:more_performance}
\end{table*}

%% file: tables/table3_1.tex
\begin{table}[t]
  \small
  \centering
  \caption{Ablation study on SEPIT's architecture.}
   \vspace{-4pt}
    \label{tab:ablation}
    \begin{adjustbox}{width=0.475\textwidth}
    \begin{tabular}{lcccc}
    \toprule
    \multirow{2}{*}{\textbf{Model}}   & \multicolumn{3}{c}{\textbf{Open-Ended}} & \textbf{Closed-Set} \\ \cmidrule(lr){2-4} \cmidrule(lr){5-5} 
                             & \textbf{BLEU-2}   & \textbf{ROUGE-L}  & \textbf{METEOR}  & \textbf{Accuracy}          \\ \midrule
    SEPIT-TinyLlama-MoEs & \textbf{60.28}    & \textbf{71.13}    & \textbf{68.27}    & \textbf{79.73\%}           \\ \midrule
    w/o Structure            & $\downarrow 4.08\%$   & $\downarrow 2.81\%$   & $\downarrow 2.98\%$   & $\downarrow 1.48\%$            \\ \midrule
    w/o MoEs                 & $\downarrow 3.17\%$   & $\downarrow 2.90\%$   & $\downarrow 0.52\%$  & $\downarrow 0.86\%$            \\ \midrule
    w/o Structure \& MoEs    & $\downarrow 4.26\%$   & $\downarrow 4.07\%$   & $\downarrow 3.13\%$   & $\downarrow 4.88\%$            \\ \midrule
    w/o Stage 0              & \multicolumn{4}{c}{---}                                      \\ \midrule
    w/o SEPIT                & $\downarrow 17.83\%$  & $\downarrow 14.17\%$  & $\downarrow 12.28\%$  & $\downarrow 7.61\%$            \\ \midrule \midrule
    w/  TrEMBL                & $\downarrow 2.69\%$  & $\downarrow 2.13\%$  & $\downarrow 2.00\%$  & $\downarrow 0.26\%$ \\ \bottomrule 
    \end{tabular}
    \end{adjustbox}
     \vspace{-8pt}
\end{table}

%% file: tables/table3_2.tex
\begin{table}[t]
    \small
    \label{tab:my_label}
    \centering
    \caption{Performance of SEPIT's encoder.}
     \vspace{-4pt}
    \label{tab:ecgo}
    \begin{adjustbox}{width=0.38\textwidth}
    \begin{tabular}{lcccc}
    \toprule
    \multirow{2}{*}{\bf{Model}} & \multirow{2}{*}{\bf{EC}} & \multicolumn{3}{c}{\bf{GO}} \\
    \cmidrule(lr){3-5}
    & & \bf{BP} & \bf{MF} & \bf{CC} \\
    \midrule
    ProtBert~\citep{ProteinBERT} & 0.838 & 0.279 & 0.456 & 0.408 \\
    OntoProtein~\citep{OntoProtein} & 0.841 & 0.436 & 0.631 & 0.441 \\
    ESM1b~\citep{ESM1b} & 0.869 & 0.452 & 0.659 & 0.477 \\
    ESM2~\citep{ESM2} & 0.874 & 0.472 & 0.662 & 0.472 \\
    CDConv~\citep{CDConv} & 0.820 & 0.453 & 0.654 & 0.479 \\
    GearNet~\citep{GearNet} & 0.810 & 0.400 & 0.581 & 0.430 \\
    ProtST-ESM2~\citep{Protst} & 0.878 & \textbf{0.482} & 0.668 & 0.487 \\
    \textbf{SEPIT's Encoder} & \textbf{0.893} & 0.476 & \textbf{0.674} & \textbf{0.497} \\
    \bottomrule
    \end{tabular}
    \end{adjustbox}
     \vspace{-8pt}
\end{table}

%% file: tables/table4_1.tex
\begin{table}[t]
  \small
  \setlength\tabcolsep{1.3mm}
  \caption{Performance comparisons with different protein input formats (all models are based on TinyLlama).}
   \vspace{-4pt}
  \label{tab:consistency}
  \centering
  \begin{adjustbox}{width=0.475\textwidth}
    \begin{tabular}{l|cccccc}
    \toprule
      \textbf{Model} &
      \textbf{PIT} &
      \textbf{PIT} &
      \textbf{SEPIT} &
      \textbf{SEPIT} &
      \textbf{SEPIT} &
      \textbf{SEPIT} \\ 
      &
      &
      \textbf{-MoEs} &
      &
      &
      \textbf{-MoEs} &
      \textbf{-MoEs} \\ \midrule
    \multicolumn{7}{c}{\textbf{Input Formation}} \\ \midrule
    \textbf{Train w/ Struct.} &
      \textcolor{myred}{\ding{55}} &
      \textcolor{myred}{\ding{55}} &
      \textcolor{mygreen}{\ding{51}} &
      \textcolor{mygreen}{\ding{51}} &
      \textcolor{mygreen}{\ding{51}} &
      \textcolor{mygreen}{\ding{51}} \\ 
    \textbf{Infer w/ Struct.} &
      \textcolor{myred}{\ding{55}} &
      \textcolor{myred}{\ding{55}} &
      \textcolor{mygreen}{\ding{51}} &
      \textcolor{myred}{\ding{55}} &
      \textcolor{mygreen}{\ding{51}} &
      \textcolor{myred}{\ding{55}} \\ \midrule
    \multicolumn{7}{c}{\textbf{Open-Ended}} \\ \midrule
    \textbf{BLEU-2} &
      57.82 &
      57.92 &
      58.43 &
      57.95 &
      60.28 &
      59.98 \\ 
    \textbf{ROUGE-L} &
      68.35 &
      69.19 &
      69.13 &
      68.75 &
      71.13 &
      70.87 \\ 
    \textbf{METEOR} &
      66.19 &
      66.29 &
      67.91 &
      67.54 &
      68.27 &
      68.00 \\ 
    \textbf{BERT-F1} &
      95.26 &
      95.29 &
      95.44 &
      95.38 &
      95.64 &
      95.59 \\ \midrule
    \multicolumn{7}{c}{\textbf{Closed-Set}} \\ \midrule
    \textbf{Accuracy} &
      76.02\% &
      78.56\% &
      79.05\% &
      77.80\% &
      79.73\% &
      79.53\% \\ \bottomrule
    \end{tabular}
    \end{adjustbox}
     \vspace{-8pt}
\end{table}

%% file: tables/table4_2.tex
\begin{table}[t]
  \small
  \setlength\tabcolsep{1.3mm}
  \renewcommand{\arraystretch}{1.05}
  \caption{Performance comparisons on OOD proteins.}
   \vspace{-4pt}
  \label{tab:model_comparison_ood}
  \centering
  \begin{adjustbox}{width=0.475\textwidth}
    \begin{tabular}{lcccc}
    \toprule
    \textbf{Model} & \textbf{BLEU-2} & \textbf{ROUGE-L} & \textbf{METEOR} & \textbf{BERT-F1} \\ \midrule
    TinyLlama & 22.23 & 35.39 & 35.27 & 90.26 \\ \midrule
    Llama2 & 28.51 & 41.63 & 41.58 & 91.27 \\ \midrule
    PIT-TinyLlama & 37.59 & 50.60 & 48.61 & 92.73 \\ \midrule
    PIT-TinyLlama-MoEs & 38.37 & 50.09 & 51.10 & 92.70 \\ \midrule
    SEPIT-TinyLlama & 40.25 & 52.03 & 52.18 & 93.02 \\ 
    \ \ \textit{+ w/o input structure} & 39.75 & 51.84 & 52.24 & 93.00 \\ \midrule
    SEPIT-TinyLlama-MoEs & 41.60 & 54.26 & 51.91 & 93.30 \\
    \ \ \textit{+ w/o input structure} & 41.26 & 53.85 & 51.52 & 93.23 \\
    \bottomrule
    \end{tabular}
  \end{adjustbox}
   \vspace{-8pt}
\end{table}

%% file: tables/table5.tex
\begin{table*}[t]
  \small
  \setlength\tabcolsep{1.3mm}
  \vspace{-10pt}
  \caption{Case studies on general-purpose protein understanding ability of SEPIT.}
  \label{tab:case}
  \centering
  \begin{adjustbox}{width=\textwidth}
    \begin{tabular}{l|l}
  \toprule
  \begin{tabular}[c]{@{}l@{}} \includegraphics[width=2.75cm]{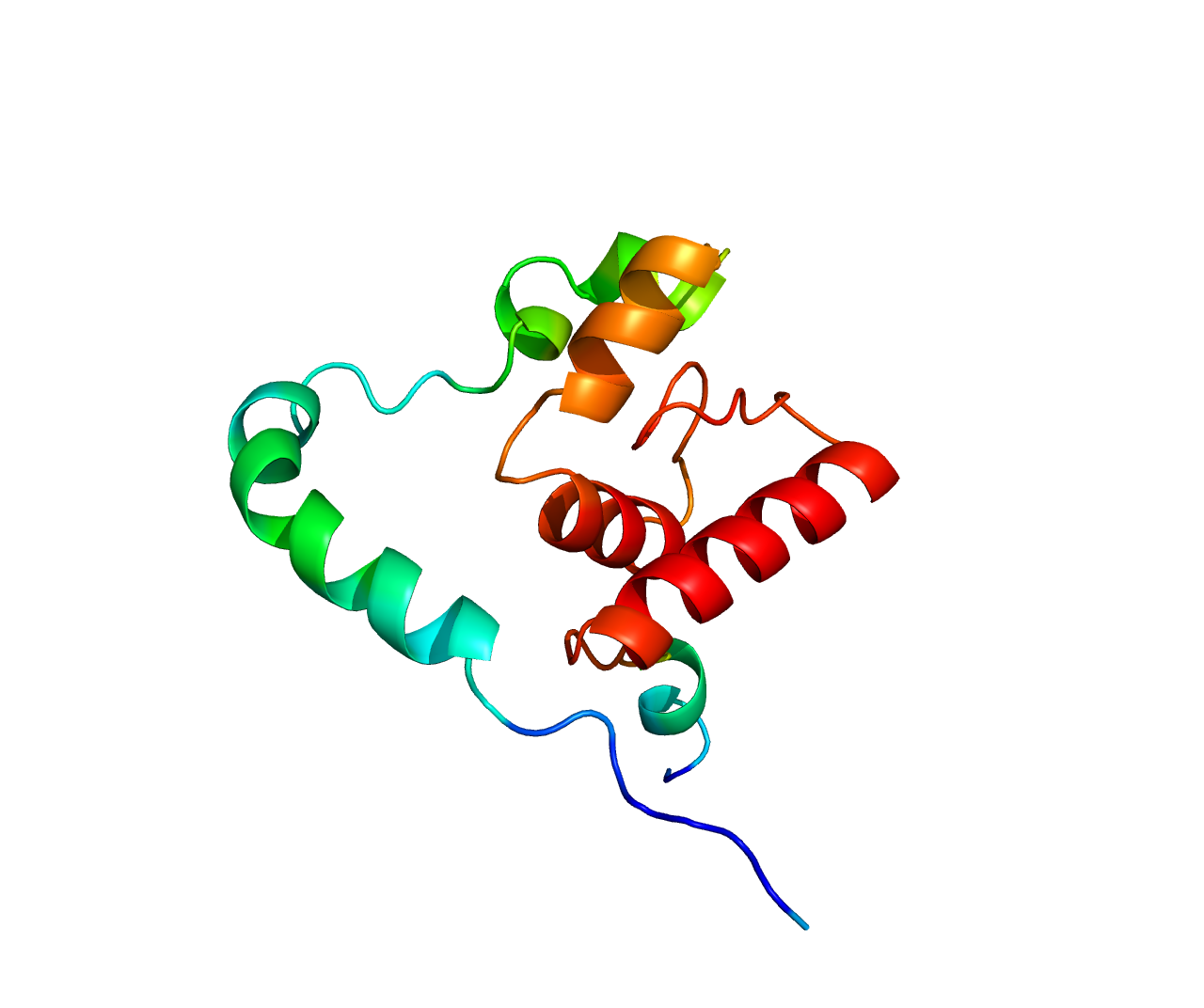} \\ P9WF36 (Swiss-Prot) \end{tabular}&
    \begin{tabular}[c]{@{}l@{}}\textbf{Question:} Explain the function of <protein>.\\ \textbf{Ground Truth:} Acts as a transcriptional regulator. Probably redox-responsive. The apo- but not holo-form probably binds DNA (By similarity).\\ \textbf{SEPIT-TinyLlama-MoEs:} \textcolor{mygreen}{Acts as a transcriptional regulator. Probably redox-responsive. The apo- but not holo-form probably binds DNA.}\\ \textbf{PIT-TinyLlama-MoEs:} \textcolor{mygreen}{Acts as a transcriptional regulator}, \textcolor{myred}{probably involved in antibiotic resistance gene expression}. \textcolor{mygreen}{Binds DNA (By similarity)}.\\ \textbf{Llama-Chat:} \textcolor{myred}{Involved in resistance toward heavy metals (PubMed:12839546). Binds cadmium or lead ions but not zinc ions (By similarity).} \\                                                \textcolor{myred}{May be involved in intracellular transport of Cd(+) or Pb(+), as well as their detoxification by sequestration into vacuoles (Probable) (PubMed:17081065)}.\\ \textbf{GPT-4-turbo:} The string you've provided appears to be an amino acid sequence of a protein. However, without additional context or data, \\                                                 I cannot provide a function caption for this specific sequence directly.\end{tabular} \\ \midrule
\begin{tabular}[c]{@{}l@{}} \includegraphics[width=2.5cm]{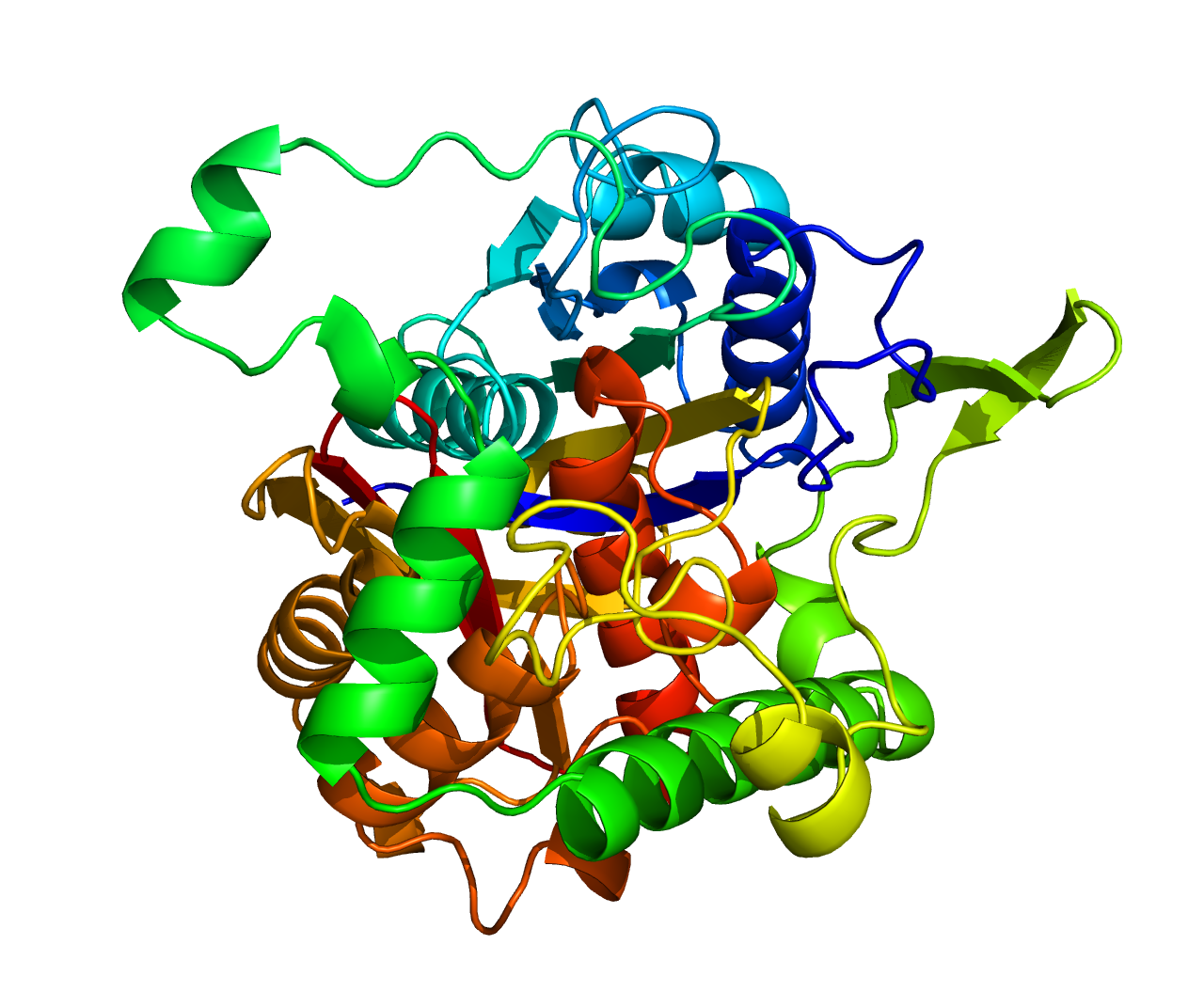} \\ A0KR49  (Swiss-Prot) \end{tabular} &
  \begin{tabular}[c]{@{}l@{}}\textbf{Question:} Which GO terms outline the biological processes of <protein>?\\ \textbf{Ground Truth:} fatty acid beta-oxidation\\ \textbf{SEPIT-TinyLlama-MoEs:} \textcolor{mygreen}{fatty acid beta-oxidation}\\ \textbf{PIT-TinyLlama-MoEs:} \textcolor{mygreen}{fatty acid beta-oxidation}; \textcolor{myblue}{phenylacetate catabolic process}\\ \textbf{Llama-Chat:} \textcolor{mygreen}{fatty acid beta-oxidation}; \textcolor{myblue}{phenylacetate catabolic process}\\ \textbf{GPT-4-turbo:} To understand the biological processes of the provided protein sequence, one must first consider its origin, functions, and structure. \\ However, without direct access to databases or running bioinformatic tools right now…\end{tabular} \\ \midrule
\begin{tabular}[c]{@{}l@{}} \includegraphics[width=2.75cm]{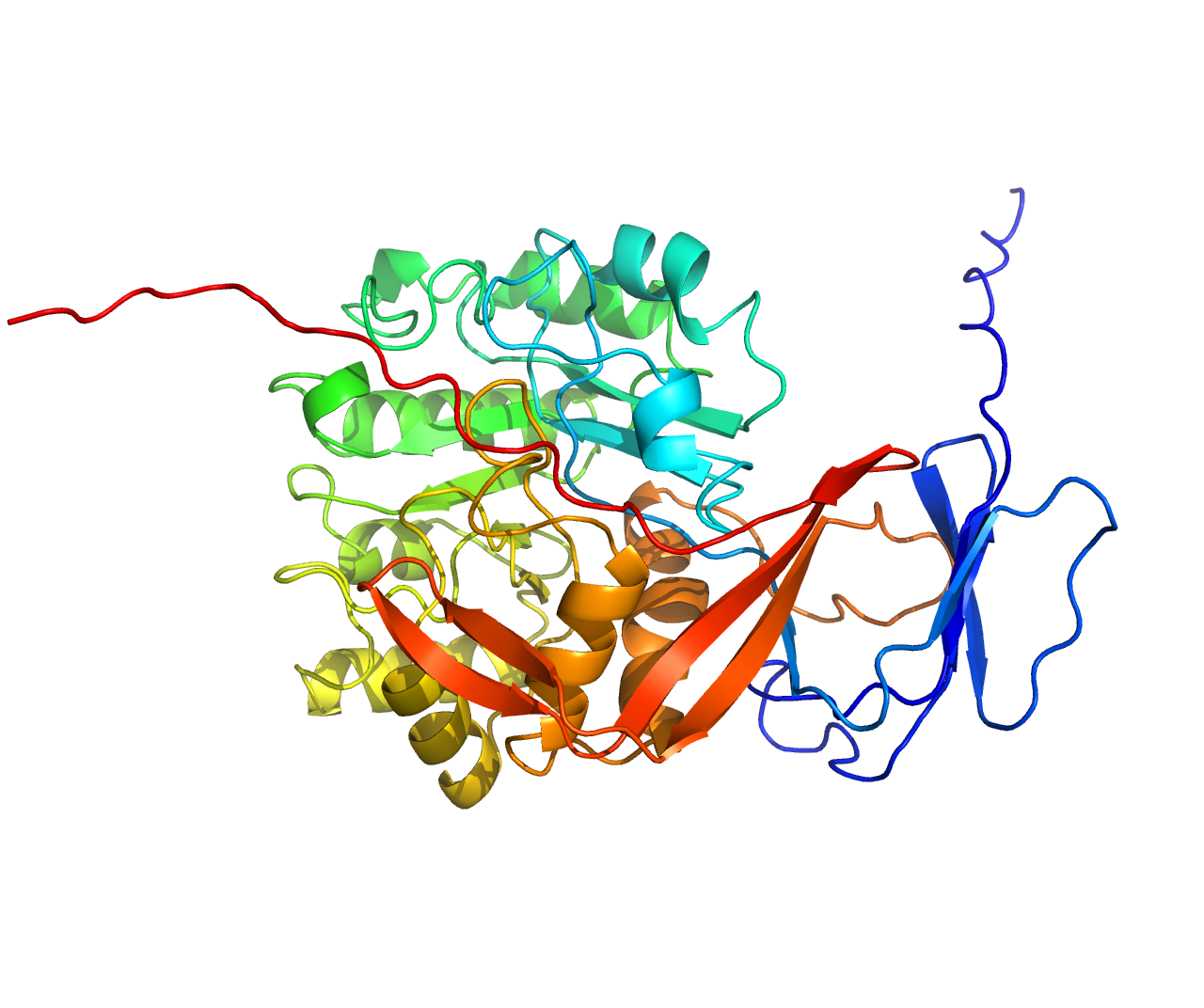} \\ Q7CS13 (Swiss-Prot) \end{tabular}  &
  \begin{tabular}[c]{@{}l@{}}\textbf{Question:} What is the protein family of <protein>?\\ Ground Truth: Belongs to the metallo-dependent hydrolases superfamily. \textcolor{myred}{Atu3266/EF\_0837 deacetylase family.}\\ \textbf{SEPIT-TinyLlama-MoEs:} \textcolor{mygreen}{Belongs to the metallo-dependent hydrolases superfamily.}\\ \textbf{PIT-TinyLlama-MoEs:} \textcolor{mygreen}{Belongs to the metallo-dependent hydrolases superfamily.} \textcolor{myred}{Type II family}.\\ \textbf{Llama-Chat:} \textcolor{mygreen}{Belongs to the metallo-dependent hydrolases superfamily}. \textcolor{myred}{HutI family}. \\ \textbf{GPT-4-turbo:} To understand the biological processes of the provided protein sequence, one must first consider its origin, functions, and structure. \\ However, without direct access to databases or running bioinformatic tools right now…\end{tabular} \\ \bottomrule
  \end{tabular}
  \end{adjustbox}
\end{table*}

%% file: sections/conclusion.tex
\section{Conclusion}
In this work, we introduce SEPIT, a novel approach for general-purpose protein understanding. SEPIT is designed to enable LLMs to interpret both the sequence and structural information of proteins, allowing them to follow instructions and generate precise insights into protein properties and functions. To achieve this, we integrate structure-aware enhancements into pre-trained pLMs and connect them to LLMs via a linear projector. The models are then trained using a two-stage instruction tuning pipeline on the largest protein instruction dataset to date, which we constructed. Experimental results demonstrate that SEPIT significantly outperforms the state-of-the-art models, highlighting its effectiveness in advancing general-purpose protein understanding.

\section*{Acknowledgment}
This work was supported in part by the National Key R\&D Program of China (Grant No.2023YFF0725001), in part by the National Natural Science Foundation of China (Grant No.92370204), in part by the Guangdong Basic and Applied Basic Research Foundation (Grant No.2023B1515120057), Education Bureau of Guangzhou Municipality.

\newpage

%% file: sections/appendix.tex
\section{Supplement to Related Work}
\label{sec:more_related}
\paragraph{\textbf{Multimodal Instruction Tuning.}}
With the emergence of Multimodal LLMs (MLLMs) such as GPT4~\citep{GPT4} and Genimi~\citep{Gemini}, MLLMs have become a focal point of research. Initially, works like CLIP~\citep{CLIP}, ALBEF~\citep{ALBEF}, VLMo~\citep{VLMo}, SimVLM~\citep{SimVLM} emphasized exploring the cross-modal alignment~\citep{tmea} between vision and language. Subsequently, based on modal alignment, Flamingo~\citep{Flamingo} and BLIP2~\citep{Blip-2} established bridges between visual encoders and LLMs using the Perceiver Resampler and the Q-Former, respectively. Following this, PaLM-E~\citep{PaLM-E} introduced "multimodal sentences" as input, injecting real-world continuous sensor data into the LLMs in the form of language tokens, thereby endowing the model with a general multi-task capability. Additionally, efforts such as InstructBLIP~\citep{Instructblip}, LLaVA~\citep{LLaVA}, MiniGPT4~\citep{Minigpt-4}, mPLUGOwl~\citep{mPLUG-Owl}, Qwen-VL~\citep{Qwen-vl}, CogVLM~\citep{Cogvlm} applied the crucial instruction tuning technique from LLMs to MLLMs, enhancing the MLLMs' ability to follow multimodal instructions. At the same time, they introduced innovations from the perspectives such as the instruction construction, training paradigms, and model design, which in turn refreshed the performance of MLLMs in a variety of visual-language downstream tasks~\citep{MLLM_survey}. In this paper, we attempt to apply this paradigm to the protein domain, investigating the potential of endowing LLMs with general-purpose protein understanding capabilities.

\paragraph{\textbf{Learning with 3D Structural Information.}}
Although pLMs pre-trained on protein sequences have been proven to be effective in numerous tasks, the protein structure is inherently a determinant of protein function~\citep{protein,protein_define}. More effectively utilizing 3D structural information can help to understand proteins more comprehensively. To capture the impact of geometric positioning and interaction relationships among residues, a class of methods that encode 3D geometric information into rotation-invariant scalars, which was then processed through graph neural networks (GNNs)~\citep{gnn,afdgcf,gnn-spatial,pog} for message passing~\citep{Protein_RL_Survey,Protein_Struct_Survey}. For example, IEConv\citep{IEConv} utilized a multi-graph to depict primary and secondary structures through covalent and hydrogen bonds and represented the tertiary structure with the spatial 3D coordinates of atoms. By blending intrinsic and extrinsic node distances and employing hierarchical pooling, it effectively perceived all three structural levels of proteins. GearNet went further by incorporating three types of directed edges (sequential edges, radius edges and k-NN edges) into the graph, capturing information at various structural levels. On this basis, CDConv~\citep{CDConv} attempted to parameterize the kernel matrices using MLP, as opposed to employing distinct kernel matrices for varying edge types, which enabled a more flexible and efficient modeling of complex interactions between residues. Additionally, another class of methods sought to prevent the loss of 3D structural information by incorporating 3D rigid transformations into the network operations. This led to the development of geometric GNNs/Transformers characterized by SE(3) invariance and equivariance~\citep{Equiformer,Equiformer_v2,SE3-Transformer,Uni-mol,Transformer-M,Geo-Transformer}. Representative examples of this approach, such as GVP~\citep{GVP} and EvoFormer~\citep{AlphaFold2}, were utilized by ESM-IF~\citep{ESM-IF}, AlphaFold2~\citep{AlphaFold2}, respectively. Furthermore, to leverage information from both evolutionary-scale protein sequences and the relatively limited protein structures, some other methods~\citep{LM-GVP,Protein_Struct_Survey} attempted to establish a connection between these two types of data.

\section{Construction of Protein Instruction Dataset}
\label{sec:data}
\begin{figure*}[h]
  \centering
  \vspace{-8pt}
  \includegraphics[width=0.765\textwidth]{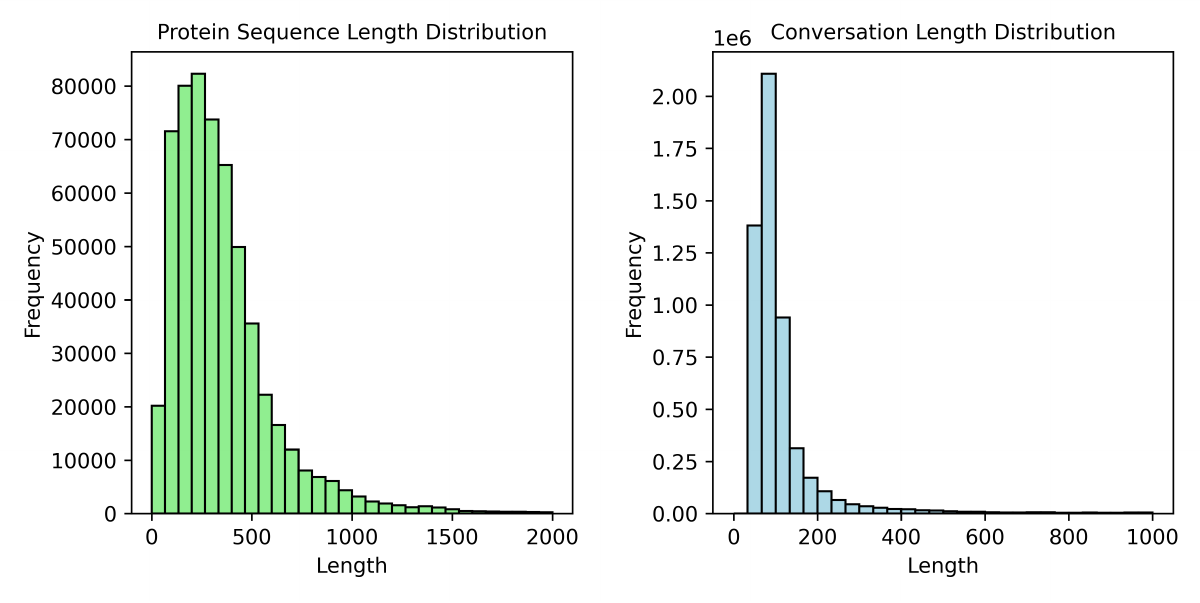}
  \vspace{-8pt}
  \caption{Distribution of protein sequence length and conversation length in the protein instruction dataset (training set).}
  \label{fig:dataset}
\end{figure*}

Currently, the widely recognized protein-text paired databases in the protein domain mainly include Swiss-Prot, TrEMBL~\citep{swiss}, and RCSB PDB~\citep{rcsbpdb}, with their specific contents detailed in Table \ref{tab:database}. Since most of the text content in TrEMBL comes from automatic annotation methods, to eliminate the impact of its unreliability on the main experiments, unless otherwise specified, this paper only uses protein-text information from Swiss-Prot and RCSB PDB databases. The corresponding structures are from AlphafoldDB~\citep{AlphafoldDB} and experimentally determined structures in RCSB PDB. The statistical information about the complete protein instruction dataset is shown in Table \ref{tab:instruction} and Figure \ref{fig:dataset}.
\begin{table*}[h]
  \small
  \setlength\tabcolsep{1.3mm}
  \caption{Protein-text paired database.}
  \vspace{-4pt}
  \label{tab:database}
  \centering
  \begin{tabular}{c|lcc}
    \toprule
    \textbf{Database} & \textbf{Content of Related Text} & \textbf{\# Protein}  & \textbf{Structure} \\
    \midrule
    Swiss-Prot & Manually calibrated structured annotations & 571,282 & AlphafoldDB \\
    \midrule
    TrEMBL & Automatic structured annotation & 248,234,451 & N/A \\
    \midrule
    RCSB PDB & Publication about protein / Meta data of protein	& 204,826 & Experimentally-determined \\
    \bottomrule
  \end{tabular}
\end{table*}

\begin{table*}[h]
  \small
  \setlength\tabcolsep{1.3mm}
  \caption{Statistical information about the protein instruction dataset.}
  \vspace{-4pt}
  \label{tab:instruction}
  \centering
  \begin{tabular}{ccccc}
\toprule
\multicolumn{1}{c|}{\textbf{Response Type}} &
  \textbf{Data Source} &
  \textbf{Question Type} &
  \textbf{\# Train Instructions} &
  \textbf{\# Test Instructions} \\ \midrule
\multicolumn{1}{c|}{\multirow{2}{*}{\begin{tabular}[c]{@{}c@{}}Open-ended\\ Generation\end{tabular}}} &
  Swiss-Prot &
  Specific property or function &
  2,487,543 &
  11,911 \\ \cmidrule{2-5} 
\multicolumn{1}{c|}{} & RCSB PDB & Protein caption & 125,703  & 2,500  \\ \midrule
\multicolumn{1}{c|}{\multirow{5}{*}{\begin{tabular}[c]{@{}c@{}}Closed-set \\ Answer\end{tabular}}} &
  RCSB PDB &
  Specific property or function &
  1,131,327 &
  22,500 \\ \cmidrule{2-5} 
\multicolumn{1}{c|}{} & GO-BP & Biological process & 601,381  & 24,162 \\ \cmidrule{2-5} 
\multicolumn{1}{c|}{} & GO-MF & Molecular function & 391,859  & 5,891  \\ \cmidrule{2-5} 
\multicolumn{1}{c|}{} & GO-CC & Cellular component & 327,780  & 6,735  \\ \cmidrule{2-5} 
\multicolumn{1}{c|}{} & EC & Enzymatic catalytic activity & 165,695  & 2,278  \\ \midrule
\multicolumn{3}{c}{\# All Instructions} & 5,231,288 & 75,977 \\ \midrule
\multicolumn{3}{c}{\# Supplemental Instructions (from TrEMBL)} & + 5,253,440 & N/A \\ \bottomrule

\end{tabular}
\end{table*}

The instruction template and example responses for Swiss-Port are as follows:
\vspace{-4pt}
\begin{itemize}[leftmargin=*, labelsep=3mm]
    \item \textbf{Function}
    \begin{itemize}
        \item What is the primary function of <protein>?
        \item What is the main function of <protein>?
        \item What is the function of <protein>?
        \item Explain the function of <protein>.
        \item What is the characteristic function associated with the protein <protein>?
        \item Can you define the function profile of the <protein>?
        \item Give me the function caption of <protein>.
        \begin{description}[leftmargin=0cm, labelindent=0cm]
            \item[Example Response] Binds to muscle nicotinic acetylcholine receptor (nAChR) and inhibit acetylcholine from binding to the receptor, thereby impairing neuromuscular transmission. Produces peripheral paralysis by blocking neuromuscular transmission at the postsynaptic site. Has a lower toxicity than cobrotoxin.
        \end{description}
    \end{itemize}
    \item \textbf{Similarity}
    \begin{itemize}
        \item Which protein family does <protein> belong to?
        \item What is the protein family of <protein>?
        \item What is the closest related protein family for <protein>?
        \item Can you identify the family or group that <protein> belongs to?
        \item To which protein family <protein> is classified?
        \item Which protein class does <protein> fall into?
        \begin{description}[leftmargin=0cm, labelindent=0cm]
            \item[Example Response] Belongs to the snake three-finger toxin family. Short-chain subfamily. Aminergic toxin sub-subfamily.
        \end{description}
    \end{itemize}
    \item \textbf{Subcellular location}
    \begin{itemize}
        \item Where is <protein> located in the cell?
        \item Can you specify the subcellular location of <protein>?
        \item What is the subcellular location of <protein>?
        \item Could you describe the subcellular location of <protein>?
        \item What are the primary subcellular regions where <protein> is detected?
        \begin{description}[leftmargin=0cm, labelindent=0cm]
            \item[Example Response] Colocalizes with ENA/VASP proteins at lamellipodia tips and focal adhesions, and F-actin at the leading edge. At the membrane surface, associates, via the PH domain, preferentially with the inositol phosphates, PtdIns(5)P and PtdIns(3)P. This binding appears to be necessary for the efficient interaction of the RA domain to Ras-GTPases (By similarity).
        \end{description}
    \end{itemize}
    \item \textbf{Induction}
    \begin{itemize}
        \item Description the effects of environmental factors of <protein>'s expression.
        \item What are the environmental factors that induce the expression of <protein>?
        \item What environmental factors causes the upregulation of <protein>?
        \item What are the environmental factors that lead to the upregulation of <protein>?
        \begin{description}[leftmargin=0cm, labelindent=0cm]
            \item[Example Response] Is slightly up-regulated when the bacterium is grown on t4LHyp or t3LHyp as sole carbon source.
        \end{description}
    \end{itemize}
    \item \textbf{Gene Ontology(Molecular Function)}
    \begin{itemize}
        \item Which GO molecular function terms have <protein> been assigned to?
        \item What molecular function is associated with <protein>?
        \item Which GO terms outline the functional capabilities of <protein>?
        \item What are the molecular functions of <protein>?
        \begin{description}[leftmargin=0cm, labelindent=0cm]
            \item[Example Response] ATP binding; protein serine kinase activity; protein serine/threonine kinase activity
        \end{description}
    \end{itemize}
    \item \textbf{Gene Ontology(Biological Process)}
    \begin{itemize}
        \item Which GO biological process terms have <protein> been assigned to?
        \item What biological process is associated with <protein>?
        \item Which GO terms outline the biological processes of <protein>?
        \item What biological processes is <protein> involved in, based on gene ontology annotations?
        \item What are the biological processes of <protein>?
        \begin{description}[leftmargin=0cm, labelindent=0cm]
            \item[Example Response] organic acid transmembrane transport; suberin biosynthetic process
        \end{description}
    \end{itemize}
    \item \textbf{Gene Ontology(Cellular Component)}
    \begin{itemize}
        \item Which GO cellular component terms have <protein> been assigned to?
        \item What cellular component is associated with <protein>?
        \item Which GO terms outline the cellular components of <protein>?
        \item What cellular components is <protein> involved in, based on gene ontology annotations?
        \item What are the cellular components of <protein>?
        \begin{description}[leftmargin=0cm, labelindent=0cm]
            \item[Example Response] cytosol; plant-type vacuole; plasma membrane
        \end{description}
    \end{itemize}
    \item \textbf{Developmental Stage}
    \begin{itemize}
        \item At which specific developmental stages is <protein> expressed?
        \item What are the developmental stages where <protein> is expressed?
        \item What are the developmental stages where <protein> is detected?
        \item What are the developmental stages where <protein> is found?
        \begin{description}[leftmargin=0cm, labelindent=0cm]
            \item[Example Response] Detected at high levels at the tube tip during early pollen germination. In germinated pollen tubes it is localized in a punctate pattern throughout the cytoplasm but most prominently at the tip region.
        \end{description}
    \end{itemize}
    \item \textbf{Short Sequence Motif}
    \begin{itemize}
        \item Can you identify and list all the motifs that are predicted to be present in <protein>?
        \item What are the short sequence motifs that are predicted to be present in <protein>?
        \item What are the short sequence motifs that are present in <protein>?
        \item What are the short sequence motifs that are found in <protein>?
        \begin{description}[leftmargin=0cm, labelindent=0cm]
            \item[Example Response] Nucleotide carrier signature motif
        \end{description}
    \end{itemize}
    \item \textbf{Tissue Specificity}
    \begin{itemize}
        \item In which tissues is the expression of <protein> absent?
        \item Describe the tissue-specific expression pattern of <protein>?
        \item What is the tissue-specific expression pattern of <protein>?
        \item What are the tissues where <protein> is expressed?
        \begin{description}[leftmargin=0cm, labelindent=0cm]
            \item[Example Response] Expressed in the ciliated cells of the airway epithelium. Not detected in the mucous cells.
        \end{description}
    \end{itemize}
    \item \textbf{Activity Regulation}
    \begin{itemize}
        \item Describe the activity regulatory mechanism of <protein> associated enzymes, transporters, microbial transcription factors.
        \item What is the activity regulatory mechanism of <protein>?
        \item Tell me about the activity regulatory mechanism of <protein>.
        \begin{description}[leftmargin=0cm, labelindent=0cm]
            \item[Example Response] Activity is sensitive to salt concentration, a high concentration of KCL (500 mM) is needed for complete inactivation.
        \end{description}
    \end{itemize}
    \item \textbf{Pathway}
    \begin{itemize}
        \item What is the role of <protein> in the metabolic pathway?
        \item Which metabolic pathway does <protein> associate with?
        \item What is the metabolic pathway that <protein> is involved in?
        \begin{description}[leftmargin=0cm, labelindent=0cm]
            \item[Example Response] Ketone degradation; acetoin degradation.
        \end{description}
    \end{itemize}
\end{itemize}

The instruction template and example responses for RCSB PDB are as follows, where \{GO\} and \{EC\} are replaced with their actual meanings (text) corresponding to GO and EC annotations.

\begin{itemize}[leftmargin=*, labelsep=3mm]
    \item \textbf{Caption}
    \begin{itemize}
        \item Tell me about this protein <protein>.
        \item Give me some information about <protein>.
        \item Give me the abstract of <protein>.
        \item Give me a comprehensive description of <protein>.
        \item Tell me about <protein>.
        \begin{description}[leftmargin=0cm, labelindent=0cm]
            \item[Example Response] The FANCM/FAAP24 heterodimer has distinct functions in protecting cells from complex DNA lesions such as interstrand crosslinks. These functions rely on the biochemical activity of FANCM/FAAP24 to recognize and bind to damaged DNA or stalled replication forks...
        \end{description}
    \end{itemize}
    \item \textbf{Others}
    \begin{itemize}
        \item Does this protein contain non-polymer entities, <protein>?
        \item Does this protein contain polymer entities, <protein>?
        \item Does this protein contain DNA polymer entities, <protein>?
        \item Does this protein contain RNA polymer entities, <protein>?
        \item Does this protein contain solvent entities, <protein>?
        \item Does this protein contain branched entities, <protein>?
        \item Does this protein have unmodeled polymer monomers, <protein>?
        \item Does this protein have hybrid nucleic acid polymer entities, <protein>?
        \item Does this protein have cis-peptide linkages, <protein>?
        \begin{description}[leftmargin=0cm, labelindent=0cm]
            \item[Example Response] Yes./No.
        \end{description}
    \end{itemize}
\end{itemize}

The instruction template and example responses for other closed-set
answer tasks are as follows:
\begin{itemize}[leftmargin=*, labelsep=3mm]
    \item \textbf{EC}
    \begin{itemize}
        \item Does <protein> associate with enzyme classification "\{EC\}"?
        \item Does EC term "\{EC\}" outline the enzyme classifications of <protein>?
        \item Is <protein> involved in enzyme classification "\{EC\}"?
         \begin{description}[leftmargin=0cm, labelindent=0cm]
            \item[Example Response] Yes./No.
        \end{description}
    \end{itemize}
    \item \textbf{GO-BP}
    \begin{itemize}
        \item Does <protein> associate with biological process "\{GO\}"?
        \item Does GO term "\{GO\}" outline the biological processes of <protein>?
        \item Is <protein> involved in biological process "{GO}"?
         \begin{description}[leftmargin=0cm, labelindent=0cm]
            \item[Example Response] Yes./No.
        \end{description}
    \end{itemize}
    \item \textbf{GO-CC}
    \begin{itemize}
        \item Does <protein> associate with cellular component "\{GO\}"?
        \item Does GO term "\{GO\}" outline the cellular components of <protein>?
        \item Is <protein> involved in cellular component "\{GO\}"?
         \begin{description}[leftmargin=0cm, labelindent=0cm]
            \item[Example Response] Yes./No.
        \end{description}
    \end{itemize}
    \item \textbf{GO-MF}
    \begin{itemize}
        \item Does <protein> associate with molecular function "\{GO\}"?
        \item Does GO term "\{GO\}" outline the functional capabilities of <protein>?
        \item Does <pro tein> have molecular function "\{GO\}"?
         \begin{description}[leftmargin=0cm, labelindent=0cm]
            \item[Example Response] Yes./No.
        \end{description}
    \end{itemize}
\end{itemize}

\newpage
\section{Supplement to the Experiments}
\label{sec:more_exp}
\subsection{Implementation Details}
\label{sec:more_imple}
As depicted in Figure \ref{fig:pipe}, we extensively utilize group learning rates throughout the entire training pipeline of SEPIT. We tend to assign higher learning rates to randomly initialized parameters, while opting for lower learning rates for pre-trained parameters in order to mitigate forgetting, setting the ratio between lower and higher learning rates at 0.1. In Stage 0, we actually employ ESM2-650M~\citep{ESM2} as the pLM and PubMedBert~\citep{PubMedBert} as the text encoder (Bio-BERT) to better encode biomedical text. We set the number of Gaussian Basis Kernels to 128. In Stage 1, we choose the representation from the penultimate layer of the protein encoder as input to LLMs to minimize the discrepancy between pre-training tasks and the current task. For the LLMs, we opt for TinyLlama-1.1B~\citep{TinyLlama}. In Stage 2, we continue most of the settings from Stage 1, while setting the number of experts to 4, with Top-1 expert being activated at a time. At this stage, our protein encoder is frozen. Regarding the hyper-parameter settings for SEPIT-TinyLlama-MoEs, we set higher learning rate to $5e^{-5}$ and trained for 5 epochs in Stage 0. In Stage 1, we set higher learning rate to $2e^{-5}$ and trained for 1 epoch. In Stage 2, we set higher learning rate to $5e^{-5}$ and trained for 1 epoch. For all stages, we trained on 32 Tesla V100 GPUs, with a batch size per GPU set to 4, employing a warm-up and linear decay learning rate scheduler, and set the warm-up ratio to 0.06. In all experiments, we employ AMP, Zero Optimizer~\citep{Deepspeed} based on AdamW~\citep{AdamW} and gradient checkpointing. For the hyper-parameter settings of other models, see Table \ref{tab:hyper}. For all MoE models, we implement them based on DeepSpeed-MoE, employing expert parallelism during the training process and setting the expert parallel size to 4.

For the API-based models, we have tallied the number of tokens consumed in the experiments conducted for this paper. It is worth noting that due to the high cost associated with GPT-4 API requests, we randomly sampled 5\% of the examples from the test set for testing. The specific token consumption is shown in Table \ref{tab:tokens}.
For all other models, we present the training hyper-parameter settings in Table \ref{tab:hyper}, their training costs in Table \ref{tab:train_cost}, and their inference costs in Table \ref{tab:infer_cost}.

\begin{table}[h]
  \small
  \setlength\tabcolsep{1.3mm}
  \caption{Tokens consumed by API-based models.}
  \label{tab:tokens}
  \centering
  \begin{tabular}{l|c}
    \toprule
    \textbf{API Model}      & \textbf{Token Consumption} \\ \midrule
    GPT-3.5-turbo  & $\sim$19M         \\ \midrule
    Claude-3-haiku & $\sim$19M         \\ \midrule
    GPT-4-turbo    & $\sim$0.95M *       \\ \midrule
    GPT-4o-mini    & $\sim$19M        \\ \midrule
    GPT-4o    & $\sim$0.95M *       \\ \midrule 
    DeepSeek-V3    & $\sim$19M       \\ \bottomrule
    \end{tabular}
\end{table}

\begin{table*}[h]
  \small
  \setlength\tabcolsep{1.3mm}
  \caption{Hyper-parameters of all models.}
  \label{tab:hyper}
  \centering
  \begin{adjustbox}{width=1\textwidth}
  \begin{tabular}{l|ccccccc}
\toprule
\textbf{Trained Model} &
  \textbf{Epochs} &
  \textbf{\begin{tabular}[c]{@{}c@{}}Wram-up\\ Ratio\end{tabular}} &
  \textbf{\begin{tabular}[c]{@{}c@{}}Batch Size\\ per GPU\end{tabular}} &
  \textbf{\begin{tabular}[c]{@{}c@{}}Global\\ Batch Size\end{tabular}} &
  \textbf{\begin{tabular}[c]{@{}c@{}}(Higher)\\ Learning Rate\end{tabular}} &
  \textbf{\begin{tabular}[c]{@{}c@{}}Auxiliary Loss \\ Coefficient\end{tabular}} &
  \textbf{\begin{tabular}[c]{@{}c@{}}Optimizer \\ Stage\end{tabular}} \\ \midrule
    TinyLlama-Chat               & 1 & 0.06 & 8 & 256 & $2e^{-5}$ & N/A  & Zero 1 \\ \midrule
    OpenLlama-v2                 & 1 & 0.06 & 8 & 512 & $4e^{-5}$ & N/A  & Zero 3 \\ \midrule
    Llama-Chat                   & 1 & 0.06 & 6 & 192 & $2e^{-5}$ & N/A  & Zero 3 \\ \midrule \midrule
    PIT-Stage 0                  & 5 & 0.06 & 4 & 128 & $2e^{-5}$ & N/A  & Zero 3 \\ \midrule
    PIT-TinyLlama-Stage 1        & 1 & 0.06 & 4 & 128 & $5e^{-5}$ & N/A  & Zero 2 \\ \midrule
    PIT-TinyLlama-Stage 2        & 1 & 0.06 & 4 & 128 & $2e^{-5}$ & N/A  & Zero 2 \\ \midrule
    PIT-TinyLlama-MoEs-Stage 2   & 1 & 0.06 & 4 & 256 & $1e^{-4}$ & 0.01 & Zero 2 \\ \midrule \midrule
    SEPIT-Stage 0                & 5 & 0.06 & 4 & 128 & $2e^{-5}$ & N/A  & Zero 3 \\ \midrule
    SEPIT-Llama-Stage 1          & 1 & 0.06 & 2 & 128 & $5e^{-5}$ & N/A  & Zero 3 \\ \midrule
    SEPIT-TinyLlama-Stage 1      & 1 & 0.06 & 4 & 128 & $5e^{-5}$ & N/A  & Zero 2 \\ \midrule
    SEPIT-TinyLlama-Stage 2      & 1 & 0.06 & 4 & 128 & $2e^{-5}$ & N/A  & Zero 2 \\ \midrule
    SEPIT-Llama-Stage 2          & 1 & 0.06 & 2 & 128 & $2e^{-5}$ & N/A  & Zero 3 \\ \midrule
    SEPIT-TinyLlama-MoEs-Stage 2 & 1 & 0.06 & 4 & 128 & $5e^{-5}$ & 0.01 & Zero 2 \\ \bottomrule
\end{tabular}
\end{adjustbox}
\vspace{40pt}
\end{table*}

\begin{table*}[h]
  \small
  \setlength\tabcolsep{1.3mm}
  \caption{Training cost of all models.}
  \label{tab:train_cost}
  \centering
  \begin{tabular}{l|ccc}
    \toprule
    \textbf{Trained Model} & \textbf{Parameter Size} &
      \textbf{\begin{tabular}[c]{@{}c@{}}Trainable \\ Parameters\end{tabular}} &
      \textbf{\begin{tabular}[c]{@{}c@{}}GPUs Cost \\ (Hrs. $\times$ \# V100)\end{tabular}} \\ \midrule
    TinyLlama-Chat                    & 1.1B        & 1.1B        & 44 $\times$ 32  \\ \midrule
    OpenLlama-v2                      & 3B          & 3B          & 45 $\times$ 64  \\ \midrule
    Llama-Chat                        & 7B          & 7B          & 170 $\times$ 32 \\ \midrule \midrule
    PIT-Stage 0                       & 650M + 110M & 650M        & 20 $\times$ 32  \\ \midrule
    PIT-TinyLlama-Stage 1        & 1.1B + 650M & 1.1B + 650M & 20 $\times$ 32  \\ \midrule
    PIT-TinyLlama-Stage 2        & 1.1B + 650M & 1.1B        & 50 $\times$ 32  \\ \midrule
    PIT-TinyLlama-MoEs-Stage 2   & 3.2B + 650M & 3.2B        & 68 $\times$ 64  \\ \midrule \midrule
    SEPIT-Stage 0                     & 650M + 110M & 650M        & 26 $\times$ 32  \\ \midrule
    SEPIT-Llama-Stage 1          & 7B + 650M   & 7B          & 82 $\times$ 64  \\ \midrule
    SEPIT-TinyLlama-Stage 1      & 1.1B + 650M & 1.1B        & 30 $\times$ 32  \\ \midrule
    SEPIT-TinyLlama-Stage 2      & 1.1B + 650M & 1.1B        & 50 $\times$ 32  \\ \midrule
    SEPIT-Llama-Stage 2          & 7B + 650M   & 7B          & 220 $\times$ 64 \\ \midrule
    SEPIT-TinyLlama-MoEs-Stage 2 & 3.2B + 650M & 3.2B        & 126 $\times$ 32 \\ \bottomrule
    \end{tabular}
    \vspace{30pt}
\end{table*}

\begin{table}[h]
  \small
  \setlength\tabcolsep{1.3mm}
  \caption{Inference cost of main models.}
  \label{tab:infer_cost}
  \centering
    \begin{tabular}{l|ccc}
    \toprule
    \textbf{Inferenced Model} &
      \textbf{Parameter Size} &
      \textbf{\begin{tabular}[c]{@{}c@{}}Activated \\ Parameters\end{tabular}} &
      \textbf{\begin{tabular}[c]{@{}c@{}}GPUs Cost \\ (Hrs. $\times$ \# T4)\end{tabular}} \\ \midrule
    TinyLlama-Chat            & 1.1B        & 1.1B        & 1 $\times$ 8    \\ \midrule
    OpenLlama-v2              & 3B          & 3B          & 8 $\times$ 8      \\ \midrule
    Llama-Chat                & 7B          & 7B          & 11 $\times$ 8   \\ \midrule \midrule
    PIT-TinyLlama        & 1.1B + 650M & 1.1B + 650M & 1.25 $\times$ 8 \\ \midrule
    PIT-TinyLlama-MoEs   & 3.2B + 650M & 1.1B + 650M & 1.5 $\times$ 8  \\ \midrule \midrule
    SEPIT-TinyLlama      & 1.1B + 650M & 1.1B + 650M & 1.5 $\times$ 8  \\ \midrule
    SEPIT-Llama          & 7B + 650M   & 7B + 650M   & 21 $\times$ 8   \\ \midrule
    SEPIT-TinyLlama-MoEs & 3.2B + 650M & 1.1B + 650M & 1.75 $\times$ 8 \\ \bottomrule
    \end{tabular}
\end{table}

\newpage
\subsection{Metric Explanation}
\label{sec:metric}
\paragraph{\textbf{BLEU Score}} The Bilingual Evaluation Understudy Score~\citep{BLEU} (BLEU score) is a metric used to evaluate the quality of machine-translated text against human-translated reference texts, which is calculated using n-gram precision. The general formula for calculating BLEU score is as follows, where $\mathbf{BP}$ penalize overly short translations:
\begin{equation}
    p_{n}=\frac{\sum_{C \in\{\text { Candidates }\}} \sum_{\text {n-gram } \in C} \text { Count }_{\text {clip }}(\text { n-gram })}{\sum_{C^{\prime} \in\{\text { Candidates }\}} \sum_{\text { n-gram }^{\prime} \in C^{\prime}} \operatorname{Count}\left(\text { n-gram }^{\prime}\right)},
\end{equation}

\begin{equation}
    \mathbf{BP} = 
    \begin{cases}
    1 & \text{if } c > r \\
    e^{(1-r/c)} & \text{if } c \leq r
    \end{cases},
\end{equation}

\begin{equation}
    \mathbf{BLEU}=\mathbf{BP}\cdot \exp\left(\sum_{n=1}^Nw_n\log p_n\right),
\end{equation}

\begin{equation}
    \log\mathbf{BLEU}=\min(1-\frac rc,0)+\sum_{n=1}^Nw_n\log p_n.
\end{equation}

\paragraph{\textbf{ROUGE-N Score}} ROUGE-N~\citep{ROUGE} is a widely-used automatic text evaluation metric designed to compare the similarity between generated text and reference text, which can be considered an improved version of BLEU with a focus on recall rather than precision. The general formula for calculating ROUGE score is as follows:

\begin{equation}
    \mathbf{ROUGE-N}=\frac{\sum_{C \in\{\text { Candidates }\}}\sum_{\text { n-gram }\in C}\operatorname{Count}_\text{match}(\text { n-gram })}{\sum_{C^{\prime}\in\{\text { Candidates }\}}\sum_{\text { n-gram }^{\prime}\in C^{\prime}}\operatorname{Count}(\text { n-gram }^{\prime})}.
\end{equation}

\paragraph{\textbf{ROUGE-L Score}} ROUGE-L~\citep{ROUGE} computes the overlap of the longest common subsequence (LCS) between the produced text and the standard references as follows, where $X$ represents the standard answer, and $Y$ denotes the generated answer, with their respective lengths being $n$ and $m$. $\beta$ is a hyper-parameter used to adjust the focus between precision $P_{\text{lcs}}$ and recall $R_{\text{lcs}}$:
\begin{equation}
    R_{\text{lcs}}=\frac{LCS(X,Y)}m,
\end{equation}

\begin{equation}
    P_{\text{lcs}}=\frac{LCS(X,Y)}n,
\end{equation}

\begin{equation}
    \mathbf{ROUGE-L}=\frac{(1+\beta^2)R_{lcs}P_{lcs}}{R_{lcs}+\beta^2P_{lcs}}.
\end{equation}

\paragraph{\textbf{METEOR Score}} METEOR~\citep{METEOR} addresses certain inherent shortcomings of the BLEU score by taking into account both precision and recall evaluated over the entire corpus. The general formula for calculating METEOR score is as follows, where $\text{Penalty}$ is the penalty of excessive word mismatches and $\alpha$ is a hyper-parameter:
\begin{equation}
    F=\frac{(\alpha^2+1)P}{R+\alpha P},
\end{equation}

\begin{equation}
    \mathbf{METEOR}=(1-\text{Penalty})\cdot F.
\end{equation}

\paragraph{\textbf{BERT Score}} BERT score~\citep{Bertscore} is an automatic evaluation metric for text generation, which computes a similarity score for
each token in the candidate sentence with each token in the reference sentence. The general formula for calculating BERT score is as follows, where tokens of reference sentence $x$ and candidate sentence $\hat{x}$ are represented by contextual embeddings:
\begin{equation}
    R_{\mathrm{BERT}}=\frac1{|x|}\sum_{x_i\in x}\max_{\hat{x}_j\in\hat{x}}\mathbf{x}_i^\top\mathbf{\hat{x}}_j,
\end{equation}

\begin{equation}
    P_{\mathrm{BERT}}=\frac1{|\hat{x}|}\sum_{\hat{x}_j\in\hat{x}}\max_{x_i\in x}\mathbf{x}_i^\top\mathbf{\hat{x}}_j,
\end{equation}

\begin{equation}
    F_{\mathrm{BERT}}=2\frac{P_{\mathrm{BERT}}\cdot R_{\mathrm{BERT}}}{P_{\mathrm{BERT}}+R_{\mathrm{BERT}}}.
\end{equation}

\subsection{More Ablation Studies}
\label{sec:more_ablation}
For the data ablation experiments, we also tested different forms of protein inputs, with the results shown in Table \ref{tab:ablation_data}. For both forms of protein inputs, the addition of extra low-quality data does not result in performance improvement but leads to performance degradation.
\input{tables/tableX}

\vspace{-10pt}
\subsection{More Case Studies}
\label{sec:more_case}
\begin{figure*}[t]
  \centering
  \includegraphics[width=0.75\textwidth]{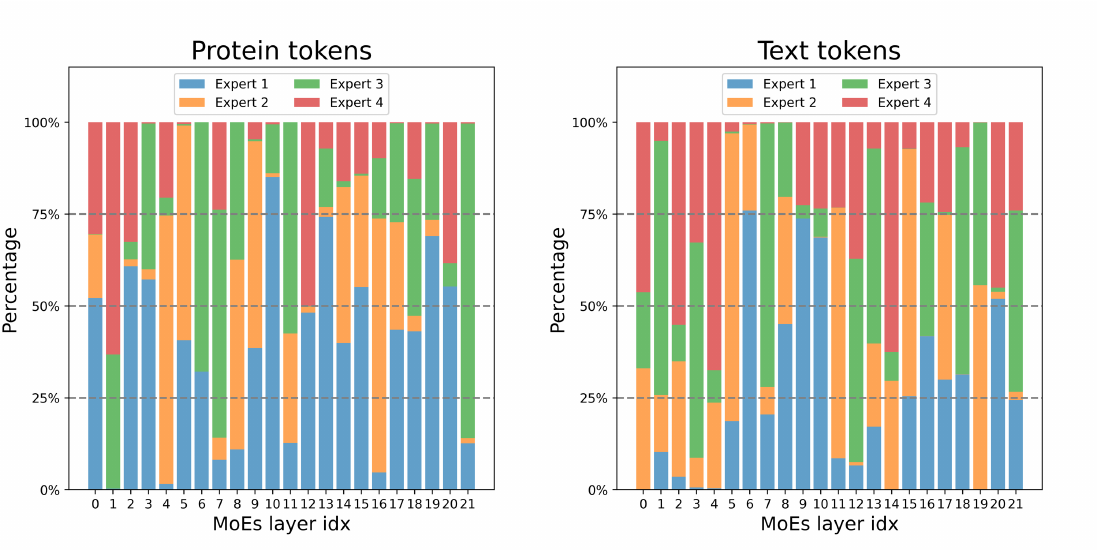}
  \caption{Workload of experts in SEPIT for protein tokens and text tokens.}
  \label{fig:moe3}
  \vspace{35pt}
\end{figure*}

\begin{figure*}[t]
  \centering
  \includegraphics[width=\textwidth]{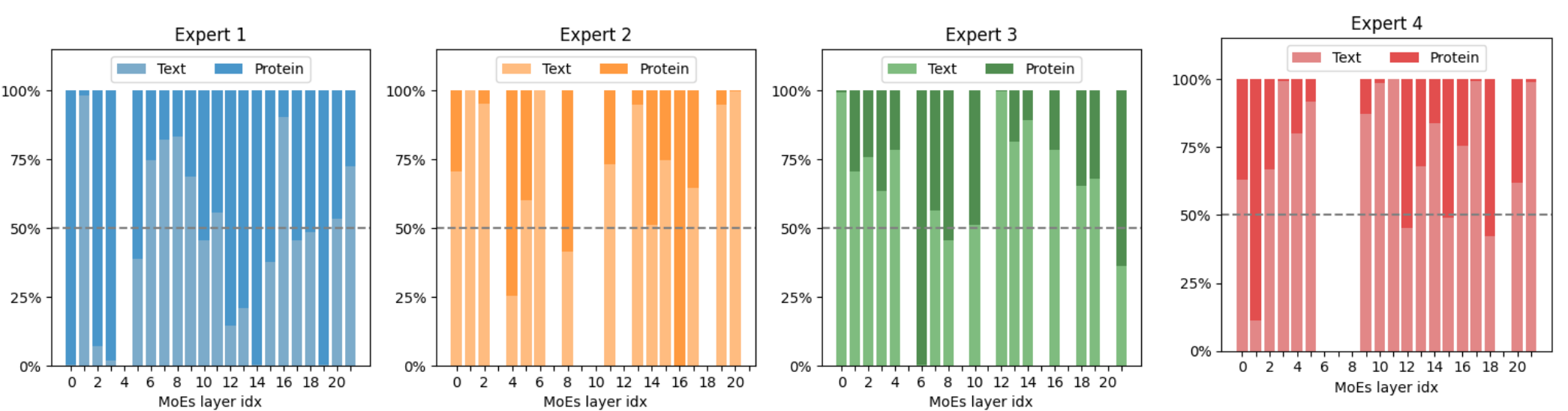}
  \caption{Workload of each expert in SEPIT for protein tokens and text tokens.}
  \label{fig:moe2}
  \vspace{35pt}
\end{figure*}

\paragraph{\textbf{More Visualization for Workload of Experts in SEPIT}}
More visualization for workload of experts in SEPIT is shown in Figure \ref{fig:moe3} and \ref{fig:moe2}. The results are similar to our analysis in the main text.

\paragraph{\textbf{More Cases for General-Purpose Protein Understanding Ability of SEPIT}}

More examples of our proposed SEPIT-TinyLlama-MoEs are shown below:

\begin{itemize}[leftmargin=*, labelsep=3mm]
    \item \textbf{O43143 (Swiss-Prot)}
    \begin{itemize}
        \item \textbf{Instruction:} What are the short sequence motifs that are present in <protein>?
        \item \textbf{Ground Truth:} DEAH box
        \item \textbf{Response:} DEAH box
    \end{itemize}
    \item \textbf{Q8DPM9 (Swiss-Prot)}
    \begin{itemize}
        \item \textbf{Instruction:} What is the function of <protein>?
        \item \textbf{Ground Truth:} Bidirectionally degrades single-stranded DNA into large acid-insoluble oligonucleotides, which are then degraded further into small acid-soluble oligonucleotides.
        \item \textbf{Response:} Bidirectionally degrades single-stranded DNA into large acid-insoluble oligonucleotides, which are then degraded further into small acid-soluble oligonucleotides.
    \end{itemize}
    \item \textbf{B4LM02 (Swiss-Prot)}
    \begin{itemize}
        \item \textbf{Instruction:} Give me the function caption of <protein>.
        \item \textbf{Ground Truth:} Plays a central role in 2-thiolation of mcm(5) S(2)U at tRNA wobble positions of tRNA(Lys), tRNA(Glu) and tRNA(Gln). Directly binds tRNAs and probably acts by catalyzing adenylation of tRNAs, an intermediate required for 2-thiolation. It is unclear whether it acts as a sulfurtransferase that transfers sulfur from thiocarboxylated URM1 onto the uridine of tRNAs at wobble position.
        \item \textbf{Response:} Plays a central role in 2-thiolation of mcm(5) S(2)U at tRNA wobble positions of tRNA(Lys), tRNA(Glu) and tRNA(Gln). May act by forming a heterodimer with ctu1/atpbd3 that ligates sulfur from thiocarboxylated urm1 onto the uridine of tRNAs at wobble position.
    \end{itemize}
    \item \textbf{Q98QW4 (Swiss-Prot)}
    \begin{itemize}
        \item \textbf{Instruction:} Which protein family does <protein> belong to?
        \item \textbf{Ground Truth:} In the N-terminal section; belongs to the phosphoglycerate kinase family.
        \item \textbf{Response:} Belongs to the phosphoglycerate kinase family.
    \end{itemize}
    \item \textbf{P0DMD4 (Swiss-Prot)}
    \begin{itemize}
        \item \textbf{Instruction:} Which protein family does <protein> belong to?
        \item \textbf{Ground Truth:} Belongs to the scolopendra neurotoxin 10 family.
        \item \textbf{Response:} Belongs to the scolopendra neurotoxin 02 (Dtx-II) family.
    \end{itemize}
    \item \textbf{A1VKS2 (Swiss-Prot)}
    \begin{itemize}
        \item \textbf{Instruction:} What is the closest related protein family for <protein>?
        \item \textbf{Ground Truth:} Belongs to the pantothenate synthetase family.
        \item \textbf{Response:} Belongs to the pantothenate synthetase family.
    \end{itemize}
    \item \textbf{P67911 (Swiss-Prot)}
    \begin{itemize}
        \item \textbf{Instruction:} What molecular function is associated with <protein>?
        \item \textbf{Ground Truth:} ADP-glyceromanno-heptose 6-epimerase activity; NADP binding
        \item \textbf{Response:} ADP-glyceromanno-heptose 6-epimerase activity; NADP binding
    \end{itemize}
    \item \textbf{Q9X5Y0 (Swiss-Prot)}
    \begin{itemize}
        \item  \textbf{Instruction:} What cellular component is associated with <protein>?
        \item \textbf{Ground Truth:} bacterial-type flagellum basal body
        \item \textbf{Response:} bacterial-type flagellum basal body
    \end{itemize}
\end{itemize}

\subsection{More Analysis}
\label{sec:more_ana}
\paragraph{\textbf{Analysis of Error Patterns}}  First, we present cases of SEPIT-TinyLlama-MoEs under major problem categories, with underlined text indicating differences from the ground truth (note that these differences are not necessarily errors). For each type, we attempt to demonstrate possible error patterns that might occur.
\\[10pt]
\noindent \textbf{Function}
\begin{itemize}[leftmargin=*, labelsep=3mm]
    \item \textbf{Instruction:} Explain the function of <protein>. (A5FQQ5)
        \begin{itemize}
            \item \textbf{Response:} This protein promotes the GTP-dependent binding of aminoacyl-tRNA to the A-site of ribosomes during protein biosynthesis.
            \item \textbf{Ground Truth:} This protein promotes the GTP-dependent binding of aminoacyl-tRNA to the A-site of ribosomes during protein biosynthesis.
        \end{itemize}
    \item \textbf{Instruction:} Give me the function caption of <protein>. (P44603)
        \begin{itemize}
            \item \textbf{Response:} Catalyzes the transfer of an acyl group from acyl-phosphate (acyl-PO(4)) to glycerol-3-phosphate (G3P) to form lysophosphatidic acid (LPA). This enzyme utilizes acyl-phosphate as fatty acyl donor, but not acyl-CoA or acyl-ACP.
            \item \textbf{Ground Truth:} Catalyzes the transfer of an acyl group from acyl-phosphate (acyl-PO(4)) to glycerol-3-phosphate (G3P) to form lysophosphatidic acid (LPA). This enzyme utilizes acyl-phosphate as fatty acyl donor, but not acyl-CoA or acyl-ACP.
        \end{itemize}
    \item \textbf{Instruction:} Can you define the function profile of the <protein>? (B0BN86)
        \begin{itemize}
            \item \textbf{Response:} Plays \uline{an essential role} in mitochondrial morphogenesis.
            \item \textbf{Ground Truth:} Plays \uline{a role} in mitochondrial morphogenesis.
        \end{itemize}
\end{itemize}

\noindent \textbf{Similarity}
\begin{itemize}[leftmargin=*, labelsep=3mm]
    \item \textbf{Instruction:} Which protein family does <protein> belong to? (Q99PZ1)
        \begin{itemize}
            \item \textbf{Response:} Belongs to the 4-hydroxy-2-oxovalerate aldolase family.
            \item \textbf{Ground Truth:} Belongs to the 4-hydroxy-2-oxovalerate aldolase family.
        \end{itemize}
    \item \textbf{Instruction:} Which protein class does <protein> fall into? (A8IQE0)
        \begin{itemize}
            \item \textbf{Response:} Belongs to the CCDC\uline{40} family.
            \item \textbf{Ground Truth:} Belongs to the CCDC\uline{39} family.
        \end{itemize}
    \item \textbf{Instruction:} To which protein family <protein> is classified? (P58600)
        \begin{itemize}
            \item \textbf{Response:} Belongs to the UDP-N-acetylglucosamine-2-epimerase family.
            \item \textbf{Ground Truth:} Belongs to the UDP-N-acetylglucosamine 2-epimerase family.
        \end{itemize}
\end{itemize}

\noindent \textbf{Subcellular location}
\begin{itemize}[leftmargin=*, labelsep=3mm]
    \item \textbf{Instruction:} Where is <protein> located in the cell? (B0KZ78)
        \begin{itemize}
            \item \textbf{Response:} Loosely associated with the inner membrane.
            \item \textbf{Ground Truth:} Loosely associated with the inner membrane.
        \end{itemize}
    \item \textbf{Instruction:} Could you describe the subcellular location of <protein>? (Q5M7N9)
        \begin{itemize}
            \item \textbf{Response:} Localizes to endoplasmic reticulum-plasma membrane contact sites (EPCS). Recruited to the cell \uline{surface} via the third C2 domain.
            \item \textbf{Ground Truth:} Localizes to endoplasmic reticulum-plasma membrane contact sites (EPCS). Recruited to the cell \uline{membrane} via the third C2 domain.
        \end{itemize}
    \item \textbf{Instruction:} What is the subcellular location of <protein>? (Q9BXH1)
        \begin{itemize}
            \item \textbf{Response:} Localized to the mitochondria in order to induce cytochrome c release.
            \item \textbf{Ground Truth:} Localized to the mitochondria in order to induce cytochrome c release.
        \end{itemize}
\end{itemize}

\noindent \textbf{Induction}
\begin{itemize}[leftmargin=*, labelsep=3mm]
    \item \textbf{Instruction:} Describe the effects of environmental factors of <protein>’s expression. (P66808)
        \begin{itemize}
            \item \textbf{Response:} By heat shock.
            \item \textbf{Ground Truth:} By heat shock \uline{at 50 degrees Celsius}.
        \end{itemize}
    \item \textbf{Instruction:} What are the environmental factors that induce the expression of <protein>? (P83404)
        \begin{itemize}
            \item \textbf{Response:} By bacterial infection.
            \item \textbf{Ground Truth:} By bacterial and \uline{parasitic hemolymph} and gut infections.
        \end{itemize}
    \item \textbf{Instruction:} What environmental factors causes the upregulation of <protein>? (D2TDA7)
        \begin{itemize}
            \item \textbf{Response:} By DNA damage, as part of the SOS response.
            \item \textbf{Ground Truth:} By DNA damage, as part of the SOS response.
        \end{itemize}
\end{itemize}

\noindent \textbf{Molecular Function/Biological Process/Cellular Component (Gene Ontology)}
\begin{itemize}[leftmargin=*, labelsep=3mm]
    \item \textbf{Instruction:} Which GO molecular function terms have <protein> been assigned to? (Q7VY99)
        \begin{itemize}
            \item \textbf{Response:} transaldolase activity
            \item \textbf{Ground Truth:} transaldolase activity
        \end{itemize}
    \item \textbf{Instruction:} What biological process is associated with <protein>? (Q6PDU1)
        \begin{itemize}
            \item \textbf{Response:} mRNA \uline{processing}; regulation of alternative mRNA splicing, via spliceosome; \uline{RNA splicing}
            \item \textbf{Ground Truth:} mRNA \uline{splicing}; regulation of alternative mRNA splicing, via spliceosome; \uline{response to vitamin E}
        \end{itemize}
    \item \textbf{Instruction:} Which GO terms outline the cellular components of <protein>? (F4HZB9)
        \begin{itemize}
            \item \textbf{Response:} plasma membrane; side of membrane
            \item \textbf{Ground Truth:} plasma membrane; side of membrane
        \end{itemize}
\end{itemize}

\noindent \textbf{Short Sequence Motif}
\begin{itemize}[leftmargin=*, labelsep=3mm]
    \item \textbf{Instruction:} Can you identify and list all the motifs that are predicted to be present in <protein>? (Q83I16)
        \begin{itemize}
            \item \textbf{Response:} 'HIGH' region; 'KMSKS' region
            \item \textbf{Ground Truth:} 'HIGH' region; 'KMSKS' region
        \end{itemize}
    \item \textbf{Instruction:} What are the short sequence motifs that are predicted to be present in <protein>? (Q5R4H0)
        \begin{itemize}
            \item \textbf{Response:} LXXLL motif; \uline{Methionine segments (Mets) motif}
            \item \textbf{Ground Truth:} LXXLL motif
        \end{itemize}
    \item \textbf{Instruction:} What are the short sequence motifs that are present in <protein>? (B7H6M7)
        \begin{itemize}
            \item \textbf{Response:} PRPP-binding
            \item \textbf{Ground Truth:} PRPP-binding
        \end{itemize}
\end{itemize}

\noindent \textbf{Tissue Specificity}
\begin{itemize}[leftmargin=*, labelsep=3mm]
    \item \textbf{Instruction:} In which tissues is the expression of <protein> absent? (Q5XIJ5)
        \begin{itemize}
            \item \textbf{Response:} Expressed in brain (at protein level).
            \item \textbf{Ground Truth:} Expressed in brain (at protein level). \uline{Expressed in hippocampal neurons}.
        \end{itemize}
    \item \textbf{Instruction:} Describe the tissue-specific expression pattern of <protein>? (Q6R2K3)
        \begin{itemize}
            \item \textbf{Response:} Expressed in roots, leaves, stems, flowers and siliques.
            \item \textbf{Ground Truth:} Expressed in \uline{seedlings}, roots, stems, leaves, flowers and siliques.
        \end{itemize}
    \item \textbf{Instruction:} What is the tissue-specific expression pattern of <protein>? (C0HJA8)
        \begin{itemize}
            \item \textbf{Response:} Expressed by \uline{the} venom duct.
            \item \textbf{Ground Truth:} Expressed by venom duct.
        \end{itemize}
\end{itemize}

\noindent \textbf{Activity Regulation}
\begin{itemize}[leftmargin=*, labelsep=3mm]
    \item \textbf{Instruction:} Describe the activity regulatory mechanism of <protein> associated enzymes, transporters, microbial transcription factors. (A3MKU1)
        \begin{itemize}
            \item \textbf{Response:} \uline{Activated} by UTP.
            \item \textbf{Ground Truth:} \uline{Inhibited} by UTP.
        \end{itemize}
    \item \textbf{Instruction:} What is the activity regulatory mechanism of <protein>? (Q9US43)
        \begin{itemize}
            \item \textbf{Response:} Allosterically activated by GTP.
            \item \textbf{Ground Truth:} Allosterically activated by GTP.
        \end{itemize}
    \item \textbf{Instruction:} Tell me about the activity regulatory mechanism of <protein>. (Q63486)
        \begin{itemize}
            \item \textbf{Response:} The activation of GTP-binding proteins is generally mediated by a guanine exchange factor (GEF), while inactivation through hydrolysis of bound GTP is catalyzed by a GTPase activating protein (GAP).
            \item \textbf{Ground Truth:} The activation of GTP-binding proteins is generally mediated by a guanine exchange factor (GEF), while inactivation through hydrolysis of bound GTP is catalyzed by a GTPase activating protein (GAP). \uline{The Ragulator complex functions as a GEF and promotes the active GTP-bound form. The GATOR1 complex functions as a GAP and stimulates RRAGA GTPase activity to turn it into its inactive GDP-bound form, preventing mTORC1 recruitment and activation}.
        \end{itemize}
\end{itemize}

\noindent \textbf{Pathway}
\begin{itemize}[leftmargin=*, labelsep=3mm]
    \item \textbf{Instruction:} What is the role of <protein> in the metabolic pathway? (Q6MUA3)
        \begin{itemize}
            \item \textbf{Response:} Pyrimidine metabolism; CTP biosynthesis via de novo pathway; CTP from UDP: step 2/2.
            \item \textbf{Ground Truth:} Pyrimidine metabolism; CTP biosynthesis via de novo pathway; CTP from UDP: step 2/2.
        \end{itemize}
    \item \textbf{Instruction:} Which metabolic pathway does <protein> associate with? (Q63AF7)
        \begin{itemize}
            \item \textbf{Response:} Carbohydrate degradation; pentose phosphate pathway; D-ribose 5-phosphate from D-ribulose 5- phosphate (non-oxidative stage): step 1/1.
            \item \textbf{Ground Truth:} Carbohydrate degradation; pentose phosphate pathway; D-ribose 5-phosphate from D-ribulose 5- phosphate (non-oxidative stage): step 1/1.
        \end{itemize}
    \item \textbf{Instruction:} What is the metabolic pathway that <protein> is involved in? (Q9SIB9)
        \begin{itemize}
            \item \textbf{Response:} Carbohydrate metabolism; tricarboxylic acid cycle; isocitrate from oxaloacetate: step 2/2.; \uline{Organic acid metabolism; propanoate degradation}.
            \item \textbf{Ground Truth:} Carbohydrate metabolism; tricarboxylic acid cycle; isocitrate from oxaloacetate: step 2/2.
        \end{itemize}
\end{itemize}

Subsequently, based on the above cases, we present the error patterns observed for each question type, with a comprehensive summary below:

\begin{figure*}[t]
  \centering
  \includegraphics[width=0.88\textwidth]{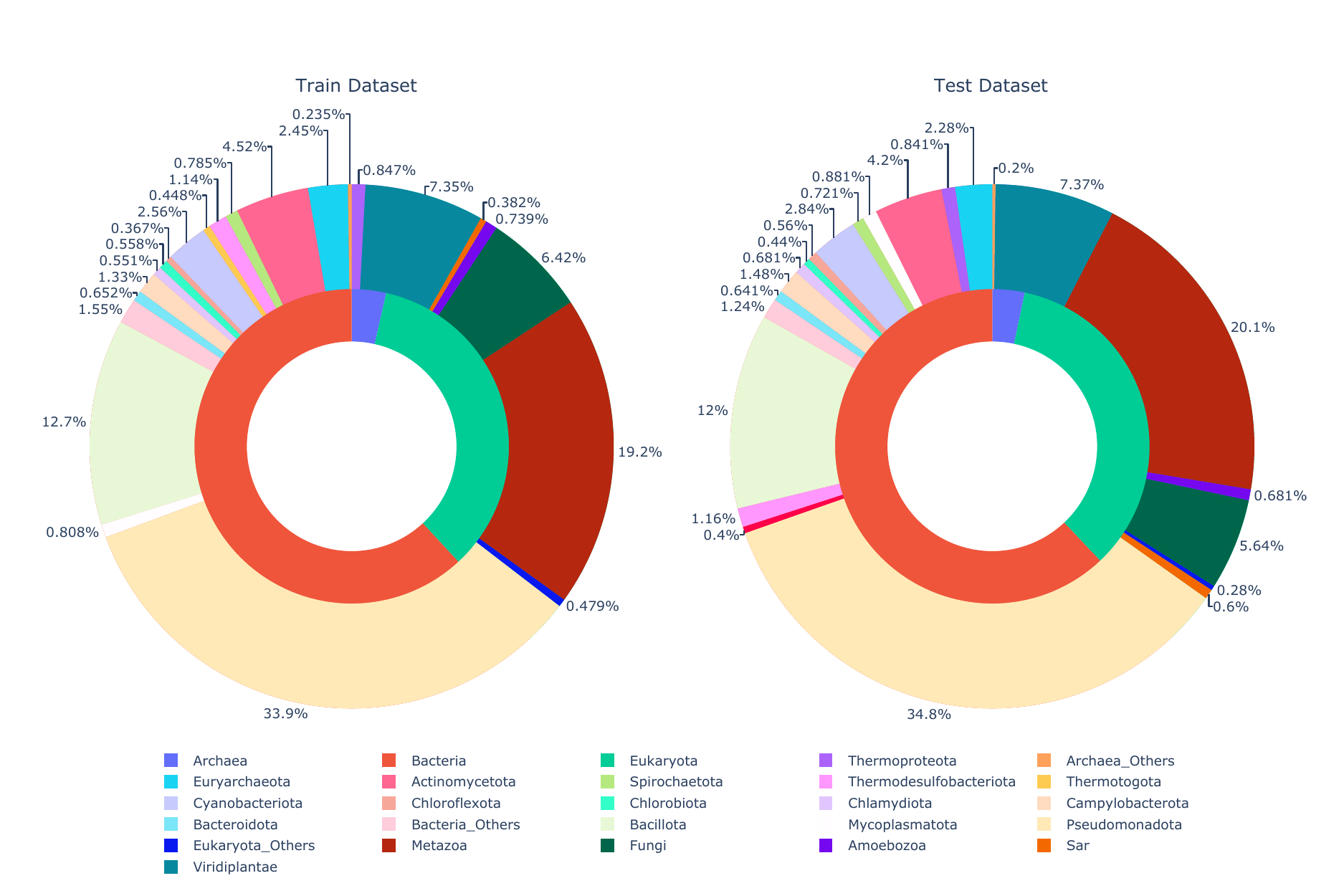}
  \vspace{-10pt}
  \caption{Taxonomic distribution at kingdom and phylum levels in open-ended generation dataset.}
  \label{fig:dis}
\end{figure*}

\begin{itemize}[leftmargin=*, labelsep=3mm]
    \item For Function, Similarity, and Subcellular Location properties and functions, our model demonstrates high prediction accuracy due to comprehensive annotation coverage and standardized formats.
    
    \item Regarding the three Gene Ontology-related properties and functions, the model achieves accurate predictions for the majority of proteins, with primary errors manifesting as either under-prediction or over-prediction of certain terms.
    
    \item For other properties and functions with adequate annotation coverage (including Induction, Short Sequence Motif, Tissue Specificity, Activity Regulation, and Pathway), the model maintains prediction accuracy, with the predominant error being term omission.
    
    \item In the case of Developmental Stage, which suffers from limited annotation coverage and poor standardization, while the model may not achieve perfect annotation accuracy, it successfully captures the essential meaning, as illustrated in this example:
    
    \begin{quote}
        \textbf{Instruction:} At which specific developmental stages is \textless protein\textgreater\ expressed? (A5HEI1)
        
        \begin{itemize}
            \item \textbf{Response:} Expressed in \textbf{embryos} (at protein level).
            \item \textbf{Ground Truth:} Detected throughout the \textbf{embryo}, covering all stages of development from pre-globular to torpedo stages.
        \end{itemize}
    \end{quote}
\end{itemize}

\begin{table}[t]
  \small
  \setlength\tabcolsep{1.3mm}
  \renewcommand{\arraystretch}{1.05}
  \caption{Performance comparison of SEPIT-TinyLlama-MoEs across different kingdoms on OOD proteins.}
  \label{tab:kingdom_comparison}
  \centering
  \begin{adjustbox}{width=0.475\textwidth}
    \begin{tabular}{lccccc}
    \toprule
    \textbf{Kingdom} & \textbf{BLEU-2} & \textbf{ROUGE-1} & \textbf{ROUGE-L} & \textbf{METEOR} & \textbf{BERT-F1} \\ \midrule
    \textit{Archaea} & 64.14 & 68.82 & 68.32 & 67.32 & 95.35 \\ \midrule
    \textit{Bacteria} & 64.73 & 72.17 & 71.16 & 69.75 & 95.69 \\ \midrule
    \textit{Eukaryota} & 52.68 & 61.84 & 59.84 & 59.75 & 94.40 \\
    \bottomrule
    \end{tabular}
  \end{adjustbox}
  \vspace{-10pt}
\end{table}

\paragraph{\textbf{Analysis Based on Taxonomic Classification}} Using the open-ended generation dataset as an example, we show the distribution of kingdom and phylum in the training and test sets in Figure~\ref{fig:dis}. Additionally, in Table~\ref{tab:kingdom_comparison}, we present the performance of SEPIT-TinyLlama-MoEs across different kingdoms on the OOD proteins (considering annotation coverage differences across kingdoms, we selected the 3 most comprehensively annotated properties and functions: Function, Similarity, and Subcellular location). Intuitively, since \textit{Eukaryota} have significantly more training data compared to \textit{Archaea}, the model would be expected to perform better on eukaryotic proteins. However, contrary to this expectation, \textit{Archaea} and \textit{Bacteria} demonstrate superior performance. We attribute this phenomenon to two primary factors:

\begin{itemize}
\item \textbf{Data complexity differences:}
\begin{itemize}
\item Although \textit{Eukaryota} has more training data, eukaryotic genomes and protein structures typically exhibit higher complexity.
\item Eukaryotic function descriptions tend to be more elaborate, detailed, and diverse, thereby increasing the complexity of the generation task.
\end{itemize}

\item \textbf{Data quality factors:} Despite the larger dataset size for \textit{Eukaryota}, it encompasses more noise and heterogeneous annotations, including a higher prevalence of technical terminology and intricate biological processes, which poses greater challenges for accurate prediction.
\end{itemize}

%% file: tables/tableX.tex
\begin{table*}[h]
  \small
  \setlength\tabcolsep{1.3mm}
  \caption{Data ablation study on SEPIT's pre-train data.}
  \label{tab:ablation_data}
  \centering
  \begin{adjustbox}{width=0.9\textwidth}
    \begin{tabular}{l|c|cccc|c}
    \toprule
    \multirow{2}{*}{\textbf{Dataset}} & \multirow{2}{*}{\textbf{Infer w/ Struct.}} & \multicolumn{4}{c|}{\textbf{Open-ended Generation}} & \textbf{Closed-set Answer} \\ \cmidrule(lr){3-6}  \cmidrule(lr){7-7} 
                                                         &   & \textbf{BLEU-2} & \textbf{ROUGE-L} & \textbf{METEOR} & \textbf{BERT-F1} & \textbf{Accuracy} \\ \midrule
    \multirow{2}{*}{Protein Instrcution Dataset (5.47M)} & \textcolor{mygreen}{\ding{51}}  & 60.28           & 71.13            & 68.27           & 95.64            & 79.73\%           \\
                                                         & \textcolor{myred}{\ding{55}}  & 59.98           & 70.87            & 68.00           & 95.59            & 79.53\%           \\ \midrule
    \multirow{2}{*}{w/ TrEMBL (+5.25M)}                  & \textcolor{mygreen}{\ding{51}}  & 58.71           & 69.65            & 66.93           & 95.42            & 79.52\%           \\
                                                         & \textcolor{myred}{\ding{55}}  & 58.55           & 69.54            & 66.78           & 95.40            & 79.48\%               \\ \bottomrule
    \end{tabular}
    \end{adjustbox}
    \vspace{30pt}
\end{table*}